\documentclass{article}

\usepackage{PRIMEarxiv}

\usepackage[utf8]{inputenc}
\usepackage[T1]{fontenc}
\usepackage{hyperref}
\usepackage{url}
\usepackage{booktabs}
\usepackage{amsfonts}
\usepackage{nicefrac}
\usepackage{microtype}
\usepackage{graphicx}
\usepackage{multirow}
\usepackage{array}
\usepackage{siunitx}
\usepackage{xcolor}
\usepackage{listings}
\usepackage{longtable}
\usepackage[noabbrev,nameinlink]{cleveref}

\usepackage{amsthm}

\usepackage{subcaption}
\usepackage{caption}
\usepackage{enumitem} 
\usepackage{algorithm}      
\usepackage{algorithmic}    
\usepackage{amssymb}

\definecolor{codegreen}{rgb}{0,0.6,0}
\definecolor{codegray}{rgb}{0.5,0.5,0.5}
\definecolor{codepurple}{rgb}{0.58,0,0.82}
\definecolor{backcolour}{rgb}{0.97,0.97,0.97}

\title{\includegraphics[width=0.99\textwidth]{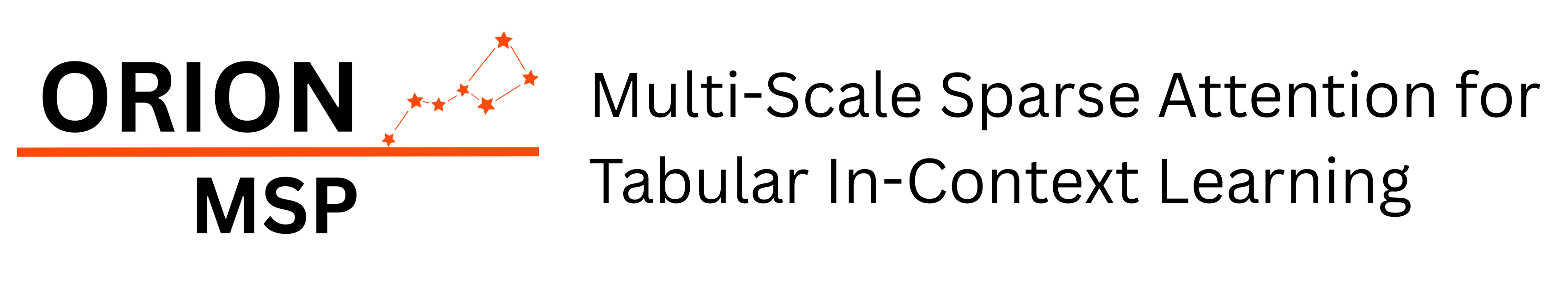}}

\author{
  Mohamed Bouadi, Pratinav Seth\\
  Aditya Tanna, Vinay Kumar Sankarapu \\
  \affiliation{Lexsi Labs, India \& France}
}

\setabstract{%
Tabular data remain the predominant format for real-world applications. Yet, developing effective neural models for tabular data remains challenging due to heterogeneous feature types and complex interactions occurring at multiple scales. Recent advances in tabular in-context learning (ICL), such as TabPFN and TabICL, have achieved state-of-the-art performance comparable to gradient-boosted trees (GBTs) without task-specific fine-tuning. However, current architectures exhibit key limitations: (1) single-scale feature processing that overlooks hierarchical dependencies, (2) dense attention with quadratic scaling in table width, and (3) strictly sequential component processing that prevents iterative representation refinement and cross-component communication. To address these challenges, we introduce Orion-MSP, a tabular ICL architecture featuring three key innovations: (1) multi-scale processing to capture hierarchical feature interactions; (2) block-sparse attention combining windowed, global, and random patterns for scalable efficiency and long-range connectivity; and (3) a Perceiver-style memory enabling safe bidirectional information flow across components. Across diverse benchmarks, Orion-MSP matches or surpasses state-of-the-art performance while scaling effectively to high-dimensional tables, establishing a new standard for efficient tabular in-context learning. The model is publicly available at \url{https://github.com/Lexsi-Labs/Orion-MSP}.
}

\setkeywords{Tabular Foundation Models, In-Context Learning, AutoML, Benchmarking.}
\runningtitle{Orion-MSP: Multi-Scale Sparse Attention for Tabular In-Context Learning}

\begin{document}
\maketitle

\section{Introduction}

Tabular data remain the most prevalent form of data in real-world applications, spanning critical systems across healthcare, finance, and scientific research. Despite the remarkable progress of deep learning in natural language processing \cite{liu2023pre,yu2024natural} and computer vision \cite{goldblum2023battle}, gradient boosted trees (GBTs) remain the predominant state-of-the-art (SOTA) for tabular prediction tasks. In other data modalities, foundation models—particularly Large Language Models (LLMs) \cite{DBLP:conf/nips/00020D24, DBLP:conf/nips/LiZLML0HY24}—have significantly advanced the ability to tackle new tasks and few-shot learning. This is largely due to their remarkable in-context learning (ICL) capabilities \cite{DBLP:conf/iclr/Zhang0YOKZY025, DBLP:conf/uss/CarliniTWJHLRBS21}, which enable them to capture patterns directly from prompts without updating their parameters. This success combined with the pervasiveness of tables have spurred interest in tabular foundation models \cite{DBLP:conf/icml/BreugelS24}.

Although LLMs are primarily designed to process natural language, recent efforts have explored fine-tuning them for tabular data tasks \cite{tabllm,tmlr}. These approaches typically rely on table serialization, which is the process of converting table rows into text or sentences suitable for tokenization. For instance, \cite{DBLP:conf/nips/0001PS24} fine-tuned a Llama 3-8B model on a large corpus of serialized tables and demonstrated that this strategy can outperform traditional tree-based models in few-shot scenarios. However, such language model–based approaches face inherent challenges. Their limited context windows restrict the number of serialized examples that can be processed simultaneously (e.g., up to 32 or 64 shots in \cite{DBLP:conf/nips/0001PS24}), and it remains uncertain whether LLMs can reliably interpret and reason over numerical values \cite{DBLP:conf/naacl/ThawaniPIS21}.

Adopting a fundamentally different strategy, the authors of \cite{tabpfn} introduced TabPFN, a transformer-based tabular foundation model designed for classification tasks and pretrained exclusively on synthetic tabular data. A key feature of TabPFN is its ability to perform in-context learning directly on tables, removing the need for tokenization and allowing efficient processing of relatively small datasets—up to 1K samples and 100 features. Building on this foundation, TabICL \cite{tabicl} introduced a simplified three-component architecture comprising: (1) column-wise embeddings via Set Transformers to capture distribution-aware feature semantics, (2) row-wise interactions with rotary positional encodings to model inter-feature dependencies, and (3) dataset-level ICL prediction through split attention, ensuring a clear separation between training and test samples. These developments position tabular foundation models as a compelling alternative to traditional approaches, particularly for zero-shot prediction tasks where dataset-specific training is infeasible.

However, current table-native ICL architectures face several fundamental limitations that hinder their practical deployment and scalability. First, existing tabular ICL architectures, including TabICL, process features uniformly at a single scale, missing hierarchical interaction patterns that naturally occur in real-world tabular data. Just as computer vision benefits from multi-scale processing—capturing edges at fine scales and objects at coarse scales—tabular data exhibits structure at multiple granularities: individual features interact locally (e.g., age and income), feature clusters form semantic groups (e.g., demographic attributes), and high-level blocks represent major data divisions (e.g., personal attributes versus behavioral patterns). Processing all features uniformly fails to capture these hierarchical relationships, limiting the model's ability to learn robust and interpretable representations.

Second, the dense attention mechanisms scale quadratically with feature count ($O(m^2)$), where $m$ denotes the number of features. While TabICL addresses sample scalability through its column-then-row architecture, the quadratic feature complexity becomes computationally prohibitive for high-dimensional tables with more than 100 features common in genomics, finance, and sensor applications. For tables with $m=100$ features, dense attention requires $10,000$ attention operations per layer, with memory requirements growing quadratically. This fundamental scalability barrier limits the practical deployment of tabular foundation models on wide real-world datasets.

Third, the strictly sequential processing pipeline in TabICL (column embedding $\rightarrow$ row interaction $\rightarrow$ ICL prediction) prevents iterative refinement and bidirectional information flow between architectural components. While each component produces rich representations, the unidirectional nature of the pipeline means that downstream insights (e.g., dataset-level patterns discovered during ICL) cannot inform upstream representations (e.g., refining feature embeddings based on dataset context). This limitation constrains the model's ability to leverage holistic dataset understanding for improved predictions, and prevents the kind of iterative refinement that has proven beneficial in multimodal architectures.

To address these limitations, we introduce \textit{Orion-MSP}, a novel tabular foundation model that extends TabICL with three synergistic architectural innovations. First, we propose multi-scale hierarchical feature processing that simultaneously captures interactions at multiple granularities (individual features, groups of 4, and groups of 16), enabling the model to learn representations at different levels of abstraction analogous to hierarchical processing in computer vision. Second, we design structured block-sparse attention patterns combining windowed local attention, global tokens for long-range dependencies, and random connectivity for universal approximation, reducing computational complexity from $O(H^2)$ to $O(H . log H)$ while maintaining expressiveness. Third, we introduce Perceiver-style cross-component memory that enables bidirectional information flow between architectural stages while provably maintaining in-context learning safety constraints—ensuring test data never influences training representations through formal ICL safety analysis.

The column-wise embedding component of Orion-MSP follows TabICL's approach \cite{tabicl}, using Set Transformers with Induced Set Attention Blocks (ISAB) \cite{isab} to create distribution-aware feature embeddings in a permutation-invariant manner. The multi-scale row interaction component processes these embeddings at multiple resolutions, with each scale using sparse attention patterns tailored to its granularity. The resulting multi-scale representations are aggregated into unified row embeddings, which then interact with the Perceiver memory before proceeding to the final ICL prediction stage. This ICL component employs split attention with label injection, ensuring proper train-test separation.

Through extensive experiments across diverse tabular benchmarks, we demonstrate that Orion-MSP achieves competitive accuracy with state-of-the-art tabular ICL methods while enabling scalability to tables with more than 100 features where existing methods fail due to memory constraints. Our work establishes that hierarchical multi-scale processing, structured sparsity, and cross-component memory can simultaneously improve both effectiveness and efficiency in tabular foundation models, opening new application domains previously inaccessible to tabular in-context learning methods.

\section{Related Work}

\textbf{Tabular In-Context Learning:} The application of in-context learning (ICL) to tabular data has recently attracted significant attention. TabPFN \cite{tabpfn} pioneered this direction by meta-training a transformer on synthetically generated datasets using structural causal models. Its encoder–decoder design allows test samples to attend to training examples, enabling zero-shot predictions without gradient-based fine-tuning. While TabPFN demonstrated strong performance on small datasets, its alternating column- and row-wise attention mechanisms make scaling to larger tables computationally prohibitive.

TabDPT \cite{tabdpt} showed that comparable performance can be achieved on real-world datasets by using similarity-based retrieval to construct contextual examples—an idea first explored in TabR \cite{tabr}. The authors extended this paradigm by integrating diffusion-based representation learning, improving robustness to missing values and distributional shifts. However, the diffusion process introduces substantial computational overhead and retains dense attention, limiting scalability. Similarly, TabPFN-v2 \cite{tabpfn2} introduced cell-based in-context learning, extending row-wise encoding to datasets exceeding $10{,}000$ samples, but it still inherits quadratic attention costs in high-dimensional tables.

Building on these foundations, TabICL \cite{tabicl} proposed a table-native transformer architecture with three components: column embedding via Set Transformers, row-wise interaction with rotary positional encodings, and an in-context learning prediction module. This design achieved state-of-the-art results across diverse benchmarks while maintaining architectural simplicity and training efficiency. Nonetheless, dense attention in row interactions and the strictly sequential pipeline limit iterative refinement, cross-component communication, and scalability to tables with more than 100 features.

ContextTab \cite{contexttab} further enhanced tabular in-context learning by incorporating contextualized feature embeddings and attention mechanisms tailored for heterogeneous tabular data. While improving performance in complex datasets, it still processes features at a single scale and relies on dense attention, limiting computational efficiency on high-dimensional tables.

Collectively, existing tabular in-context learning models demonstrate strong performance yet share core limitations: dense quadratic attention, uniform single-scale processing, and lack of cross-component feedback.

\textbf{Sparse Attention Mechanisms:} Sparse attention techniques from natural language processing offer a promising route to improve computational efficiency in tabular in-context learning. BigBird \cite{bigbird} and Longformer \cite{longformer} demonstrated that block-sparse attention patterns can approximate dense attention with linear complexity while maintaining strong theoretical guarantees. Similarly, Sparse Transformers \cite{sparse_transformers,sparse_transformers_neurips} employ structured sparsity for generative modeling, reducing computation without substantial performance degradation. Despite their success in sequential data, these methods have yet to be systematically adapted for tabular in-context learning, where the primary challenge lies in feature dimension rather than sequence length.

\textbf{Hierarchical and Multi-Scale Architectures:} Hierarchical architectures have proven effective in other domains. Funnel Transformers \cite{funnel} and Swin Transformers \cite{swin} use multi-scale processing and pooling to capture information at different resolutions, while Set Transformers \cite{settransformers,settransformers2} leverage pooling by multihead attention for permutation-invariant set processing. Although TabICL \cite{tabicl} employs Set Transformers for column embeddings, it does not incorporate hierarchical multi-scale processing or iterative pooling across feature groups, limiting its ability to model complex interactions in high-dimensional tables.

\textbf{Cross-Component Communication:} Cross-component memory and iterative refinement have shown success in multimodal learning. Perceiver \cite{perceiver} and Perceiver IO \cite{perceiverio} introduce latent bottlenecks to compress and share information across modalities, and vision-language models \cite{vision_models} leverage iterative cross-attention for refinement. However, these approaches do not address the causal constraints of in-context learning, where test examples must never influence the representation of training data, leaving a gap for tabular in-context learning.

\section{Proposed Approach: Orion-MSP}

\subsection{Problem Formulation}
Consider a tabular dataset $\mathcal{D} = \{(\mathbf{x}_i, y_i)\}_{i=1}^n$ with $n$ samples and $m$ features. Let $\mathbf{X} \in \mathbb{R}^{n \times m}$ denote the feature matrix, where each column $\mathbf{c}_j \in \mathbb{R}^n$ ($j \in \{1, \ldots, m\}$) represents the values of the $j$-th feature across all samples.
\newpage
In the in-context learning setting, we are given a \emph{context set} of $n_{\text{train}}$ labeled examples:
\begin{equation*}
    \mathcal{C} = \{(\mathbf{x}_i, y_i)\}_{i=1}^{n_{\text{train}}}
\end{equation*}
and a set of $n_{\text{test}}$ \emph{query samples}:
\begin{equation*}
    \mathcal{Q} = \{\mathbf{x}_i\}_{i=1}^{n_{\text{test}}}
\end{equation*}
Our goal is to predict the conditional distribution of the target for each query sample given the context set:
\begin{equation*}
    p(y|x, \mathcal{C}), \forall x \in \mathcal{Q}
\end{equation*}

\subsection{High-level Structure: From Data to ICL}

\begin{figure}[htbp]
    \centering
    \includegraphics[width=0.9\linewidth]{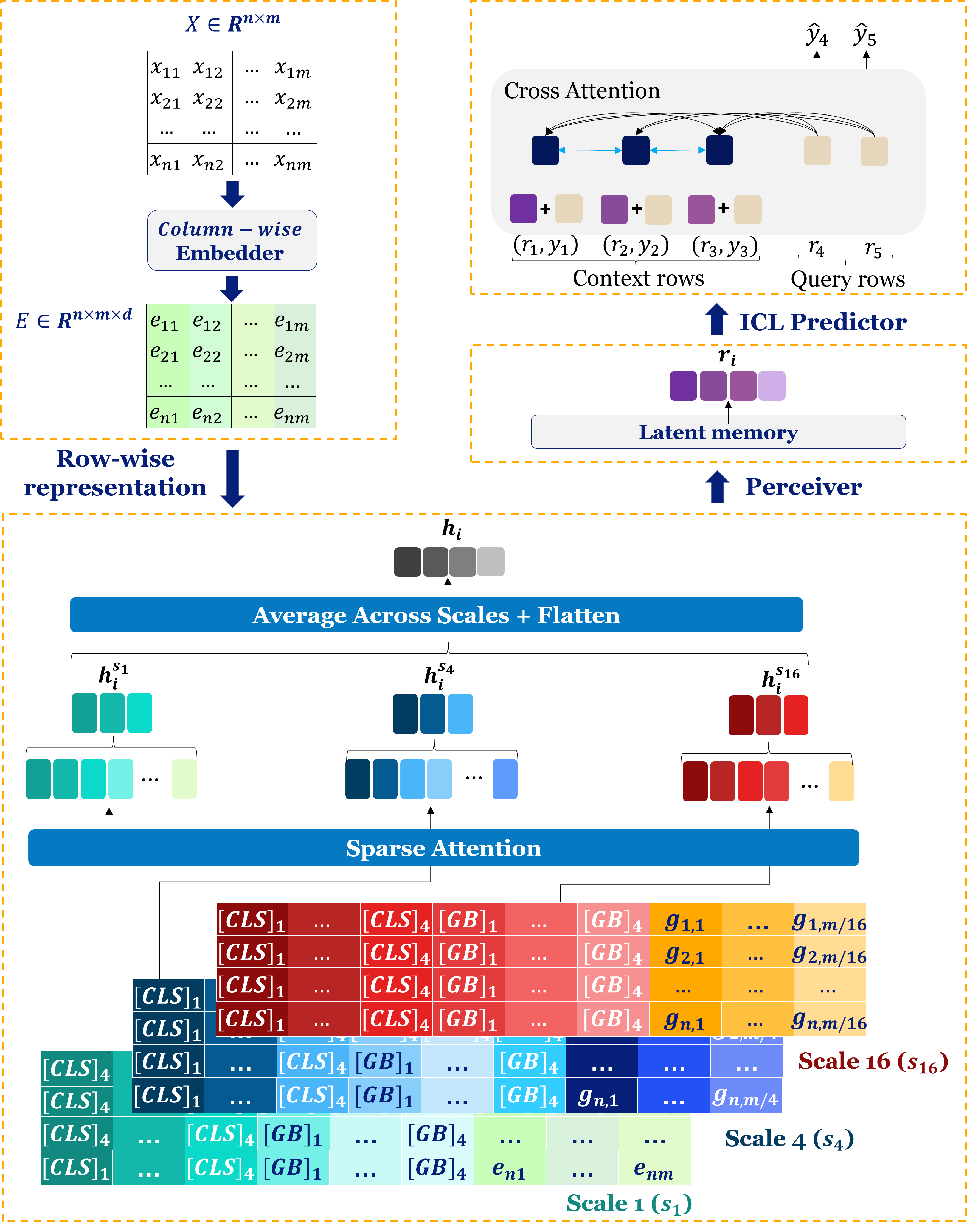}
    \caption{An overview of Orion-MSP architecture. First, column-wise embedding transforms input table into embedding vectors $E$. Next, multi-scale sparse row interaction prepends learnable \textit{[CLS]} and \textit{[GLOBAL]} tokens to $E$, processes features at multiple granularities (scales 1, 4, and 16) with sparse attention transformers, and aggregates \textit{[CLS]} outputs across scales to yield row embeddings $H$. Cross-component Perceiver memory enables bidirectional communication: training rows write to latent memory, which all rows read for enhanced representations $R$. Finally, ICL predicts test labels from $R$ in a single forward pass.}
    \label{fig:archi}
\end{figure}

Orion-MSP consists of four core components that collectively enable efficient and generalizable tabular in-context learning: (1) Column Embedding: transforms raw tabular features into dense, semantically meaningful representations; (2) Multi-Scale Sparse Row Interaction: captures dependencies at multiple granularities via an hierarchy of attention scales, combining \textit{CLS} and \textit{GLOBAL} tokens for local and long-range connectivity; (3) Cross-Component Perceiver Memory: introduces a latent memory bottleneck that enables safe bidirectional communication between modules, promoting iterative refinement without information leakage; (4) Dataset-wise In-Context Learning Predictor: leverages the enriched representations to perform zero-shot prediction across new tasks without gradient updates. An overview of the complete architecture is shown in Figure~\ref{fig:archi}.

\textit{Orion-MSP} extends the original TabICL \cite{tabicl} architecture with three complementary innovations designed to address the fundamental challenges of tabular data processing: computational inefficiency, limited feature interaction modeling, and the need for hierarchical pattern recognition. Our approach maintains the core in-context learning paradigm while introducing architectural enhancements that significantly improve both efficiency and performance.

\subsection{Column-wise Embedding}

Tabular data exhibits unique characteristics compared to other modalities: each column represents a distinct feature with its own distribution, scale, and statistical properties (e.g., mean, variance, skewness, kurtosis). To capture these distributional characteristics, we adopt the original TabICL \cite{tabicl} column-wise embedder to map each scalar cell in a column \(c_j \in \mathbb{R}^{n}\) to a \(d\)-dimensional representation using a \emph{shareable Set Transformer}, \(TF_{\text{col}}\), that treats the column as a permutation-invariant set of values. Our goal is to transform each cell value $X_{ij}$ into a $d$-dimensional embedding $\mathbf{E}_{ij} \in \mathbb{R}^d$ that encodes both:
\begin{enumerate}
    \item The value of the cell ($X_{ij}$)
    \item The distributional context of the column ($\mathbf{c}_j$)
\end{enumerate}

This differs fundamentally from standard embedding approaches (e.g., word embeddings) where each discrete token has a fixed embedding regardless of context. In tabular data, the \textit{meaning} of a value depends heavily on the column's distribution: a value of 50 may be typical in one feature but an outlier in another.

Concretely, \(TF_{\text{col}}\) predicts a per-cell affine map, assigning each cell its own weight and bias. The process consists of three main steps:

\subsubsection{Initial Projection:}
Project the column values into a $d$-dimensional embedding space:
    \begin{equation}
    \mathbf{U}_j = \text{Linear}_{\text{proj}}(\mathbf{c}_j) \in \mathbb{R}^{n \times d}
    \end{equation}
    where $\text{Linear}_{\text{proj}}: \mathbb{R} \to \mathbb{R}^d$ is a learned linear transformation. This creates initial token embeddings for each cell in the column.

\subsubsection{Induced Set Attention Blocks (ISAB):}
To efficiently capture global distributional information while maintaining computational tractability, we employ ISAB \cite{isab} with $k$ learnable inducing points. It consists of two sequential Multi-Head Attention Blocks $(\text{MAB}_{1}, \text{MAB}_{2})$:
    
\begin{align}
    \mathbf{M}_j &= \text{MAB}_{1}(\mathbf{I}, \mathbf{U}_j^{\text{train}}, \mathbf{U}_j^{\text{train}}) \in \mathbb{R}^{k \times d} \label{eq:isab_encode} \\
    \mathbf{V}_j &= \text{MAB}_{2}(\mathbf{U}_j, \mathbf{M}_j, \mathbf{M}_j) \in \mathbb{R}^{n \times d} \label{eq:isab_decode}
\end{align}
    
where $\mathbf{I} \in \mathbb{R}^{k \times d}$ denote $k$ trainable inducing point embeddings ($k \ll n$), which serve as a compressed representation of the column distribution.
    
We define a Multi-Head Attention Block as:
    
\begin{equation}
    \text{MAB}(\mathbf{Q}, \mathbf{K}, \mathbf{V}) = \text{LayerNorm}(\mathbf{H} + \text{MultiHead}(\mathbf{Q}, \mathbf{K}, \mathbf{V}))
\end{equation}
    
where $\mathbf{H}$ is a residual connection (set to $\mathbf{Q}$ if dimensions match, otherwise passed through a projection), and:
    
\begin{equation}
    \text{MultiHead}(\mathbf{Q}, \mathbf{K}, \mathbf{V}) = \text{Concat}(\text{head}_1, \ldots, \text{head}_h)\mathbf{W}_O
\end{equation}
    
with each head defined as:
\begin{equation}
    \text{head}_i = \text{Attention}(\mathbf{Q}\mathbf{W}_Q^i, \mathbf{K}\mathbf{W}_K^i, \mathbf{V}\mathbf{W}_V^i)
\end{equation}

Following TabICL \cite{tabicl}, we use \(d{=}128\), \(k{=}128\), 4 heads, and 3 ISAB blocks. Crucially, in Equation~\ref{eq:isab_encode}, we use only \textit{training samples} $\mathbf{U}_j^{\text{train}} \in \mathbb{R}^{n_{\text{train}} \times d}$ as keys and values. This ensures that the inducing points $\mathbf{M}_j$ capture the distribution of the training data only, preventing information leakage from test samples during embedding. This is crucial for maintaining the in-context learning paradigm.
    
In Equation~\ref{eq:isab_decode}, all samples (training and test) query the inducing points to obtain their contextualized embeddings. The inducing points act as a distributional summary: they encode statistical properties (e.g., mean, variance, skewness) of the training column values, and each cell embedding is adjusted based on where it lies within this learned distribution.
    
\subsubsection{Weight and Bias Generation:}
The ISAB output $\mathbf{V}_j$ is passed through a feedforward network to generate cell-specific weights and biases:

\begin{equation}
    \mathbf{W}_j, \mathbf{B}_j = \text{FFN}(\mathbf{V}_j), \mathbf{W}_j, \mathbf{B}_j \in \mathbb{R}^{n \times d}
\end{equation}
    
where:
\begin{align}
    \text{FFN}(\mathbf{V}_j) &= \text{Linear}_{\text{out}}(\text{GELU}(\text{Linear}_{\text{hidden}}(\mathbf{V}_j)))
\end{align}
    
The final embeddings are then computed as:    
\begin{equation}
    \mathbf{E}_{:,j,:} = \mathbf{W}_j \odot \mathbf{c}_j + \mathbf{B}_j \in \mathbb{R}^{n \times d}
\end{equation}
    
where $\odot$ denotes element-wise (Hadamard) product, and $\mathbf{c}_j$ is broadcasted to shape $(n, d)$.
    
This formulation allows each cell's embedding to be a \textit{function} of both its raw value ($\mathbf{c}_j$) and the column's learned distributional properties ($\mathbf{W}_j, \mathbf{B}_j$).

Note that, in our architecture, row-wise interaction requires prepending special tokens (e.g., [CLS], [GLOBAL]) to each row. To accommodate these, the column embedding reserves $C$ positions at the beginning of each column:

\begin{equation}
\mathbf{E} \in \mathbb{R}^{n \times (m + C) \times d}
\end{equation}

For the reserved positions (indices $1$ to $C$), we use a \textit{skippable linear layer} that outputs zeros or small random values:

\begin{equation}
\mathbf{E}_{:,j,:} = \begin{cases}
\text{SkipLinear}(\mathbf{c}_j) & \text{if } j \leq C \text{ (reserved)} \\
\mathbf{W}_j \odot \mathbf{c}_j + \mathbf{B}_j & \text{if } j > C \text{ (features)}
\end{cases}
\end{equation}

where $\text{SkipLinear}$ is a linear layer with very small initialization, allowing the model to learn appropriate embeddings for reserved positions during training.

The Set Transformer architecture ensures that $\text{TF}_{\text{col}}$ is \textit{permutation-invariant} with respect to the order of samples within a column. Formally, let $\pi: [T] \to [T]$ be any permutation, and let $\mathbf{c}_j' = \mathbf{P}_\pi \mathbf{c}_j$ where $\mathbf{P}_\pi$ is the corresponding permutation matrix. Then:

\begin{equation}
\text{TF}_{\text{col}}(\mathbf{c}_j') = \mathbf{P}_\pi \, \text{TF}_{\text{col}}(\mathbf{c}_j)
\end{equation}

This property is inherited from the attention mechanism in ISAB, where the softmax normalization and weighted aggregation are invariant to input order.

The inducing points $\mathbf{M}_j \in \mathbb{R}^{k \times d}$ learned by the first MAB serve as a \textit{distributional summary} of column $j$. Empirically, we observe that:
\begin{itemize}
    \item Columns with similar statistical moments (mean, variance, skewness, kurtosis) have similar inducing point representations (measured by cosine similarity).
    \item The inducing points capture multi-modal distributions: for categorical features encoded numerically, different modes correspond to different cluster centers in the inducing point space.
    \item Outliers in $\mathbf{c}_j$ receive distinct embeddings, as their attention weights to $\mathbf{M}_j$ differ significantly from typical values.
\end{itemize}

\subsection{Multi-Scale Sparse Row-Wise Interaction}
\label{sec:multiscale}

While column-wise embedding captures distributional properties of individual features, row-wise interaction must model complex dependencies \textit{across features} to extract meaningful sample representations. However, directly applying dense self-attention to all feature tokens incurs quadratic complexity $O(m^2)$ and may overfit when the number of features varies significantly across datasets. To address these challenges, we introduce a \textbf{hierarchical multi-scale sparse attention} mechanism that processes features at multiple granularities with efficient block-sparse patterns.

\subsubsection{Motivation and Design Principles}

Tabular datasets exhibit several unique characteristics that complicate feature interaction modeling:

\begin{enumerate}
    \item \textbf{Variable feature counts}: The number of features $m$ varies dramatically across datasets, making fixed-scale architectures suboptimal.
    \item \textbf{Heterogeneous feature relationships}: Some features interact locally (e.g., age and age-related health metrics), while others have global dependencies (e.g., categorical indicators).
    \item \textbf{Computational constraints}: Dense attention over $m$ features has complexity $O(m^2)$, becoming prohibitive for wide tables or long context windows.
    \item \textbf{Overfitting risks}: Full attention can memorize training-specific feature correlations that do not generalize to new datasets.
\end{enumerate}

Inspired by hierarchical representations in vision~\cite{ying2018} and multi-resolution modeling in speech~\cite{2021wavesplit}, Orion-MSP decomposes feature interactions into multiple resolution levels:

\begin{itemize}
    \item \textbf{Fine scale ($s=1$)}: Captures detailed pairwise dependencies between individual features.
    \item \textbf{Coarse scales ($s>1$)}: Aggregates semantically related features into groups, reducing sequence length and enabling broader contextual reasoning.
    \item \textbf{Scale aggregation}: Combines representations across scales to balance local precision and global context.
\end{itemize}

\begin{figure}[htbp]
    \centering
    \includegraphics[width=1\linewidth]{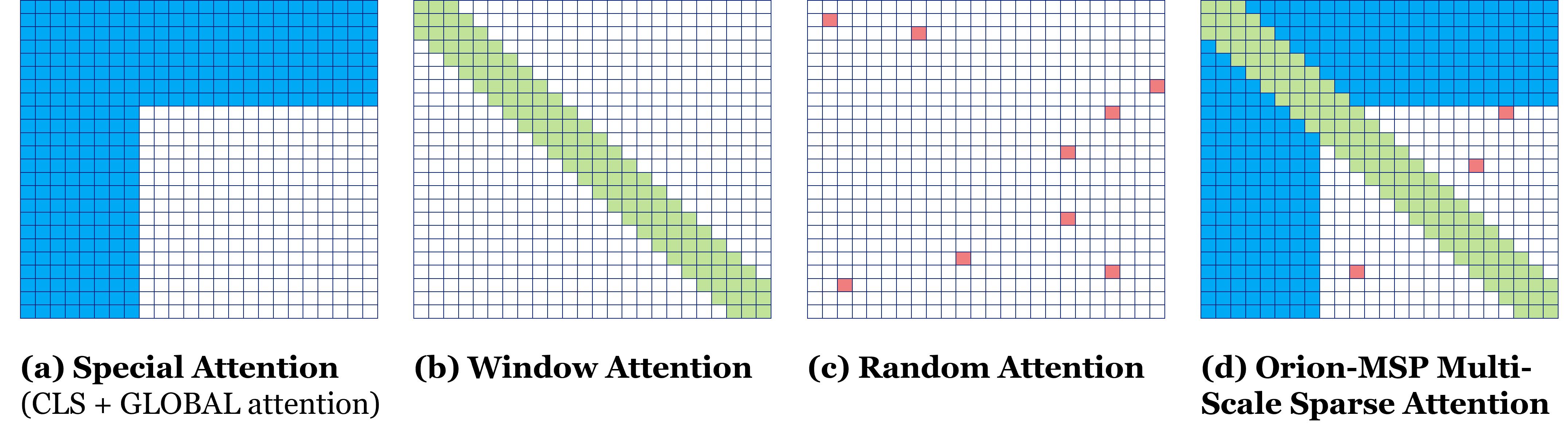}
    \caption{Building blocks of the attention mechanism used in Orion-MSP. White color indicates absence of attention. (a) special attention includes $CLS=4$ and global attention with $GB=4$, (b) sliding window attention with $w = 8$, (c) random attention with $r = 2$, (d) the combined row representation of Orion-MSP model.}
    \label{fig:sparse}
\end{figure}

To further improve efficiency and generalization, we adopt a block-sparse attention pattern inspired by Longformer~\cite{longformer} and BigBird~\cite{bigbird}, as depicted in Figure \ref{fig:sparse}:

\begin{itemize}
    \item \textbf{Sliding window attention}: Local connectivity within a fixed radius $w$, preserving fine-grained structure.
    \item \textbf{Global tokens}: Specialized tokens with full connectivity, ensuring stable long-range information flow.
    \item \textbf{Random links}: Optional sparse stochastic connections that enhance expressivity and global reachability..
\end{itemize}

This design reduces attention complexity from $O(m^2)$ to $O(m \cdot (w + g + r))$ where $w$, $g$ and $r$ are the window size, the number of global tokens, and the number of random links respectively.

Formally, the multi-scale sparse row-wise transformer, $\text{TF}_{\text{row}}^{\text{MS}}$, processes column-embedded features $\mathbf{E} \in \mathbb{R}^{B \times n \times (m+C) \times d}$ to generate row-wise embeddings $\mathbf{H} \in \mathbb{R}^{B \times n \times (N{\text{cls}} \cdot d)}$: 

\begin{equation}
\mathbf{H} = \text{TF}_{\text{row}}^{\text{MS}}(\mathbf{E}, d_{\text{valid}}) \in \mathbb{R}^{B \times n \times (N_{\text{cls}} \cdot d)}
\end{equation}

where $B$ is the number of datasets, $n$ the number of samples per dataset, $m$ the number of features, and $d$ the embedding dimension.
The constant $C = N_{\text{cls}} + N_{\text{global}}$ accounts for special token slots, and$d_{\text{valid}} \in \mathbb{R}^B$ optionally indicates the number of valid features per dataset for handling variable-length inputs.

The transformation proceeds through the following steps:

\subsubsection{Multi-Scale Feature Grouping:}

First, for each scale $s \in \mathcal{S} = \{s_1, s_2, \ldots, s_M\}$ (e.g., $\mathcal{S} = \{1, 4, 16\}$), we group the $m$ feature tokens into $K_s = \lceil m / s \rceil$ groups of size $s$.

The default grouping strategy uses a \textit{learnable soft grouping} via Pooling by Multihead Attention (PMA)~\cite{pma} to adaptively attend to features:
    
\begin{align*}
    \mathbf{Q}_s &= \text{Seed}_s + \text{PE}(K_s) \in \mathbb{R}^{K_s \times d} \\
    \mathbf{K}_s &= \text{Linear}_K(\mathbf{E}_{:,:,C:,:}) \in \mathbb{R}^{B \cdot n \times m \times d} \\
    \mathbf{V}_s &= \text{Linear}_V(\mathbf{E}_{:,:,C:,:}) \in \mathbb{R}^{B \cdot n \times m \times d} \\
    \mathbf{A}_s &= \text{softmax}\left(\frac{\mathbf{Q}_s \mathbf{K}_s^\top}{\sqrt{d}}\right) \in \mathbb{R}^{K_s \times n} \\
    \mathbf{F}_s &= \mathbf{A}_s \mathbf{V}_s \in \mathbb{R}^{B \times n \times K_s \times d}
\end{align*}
    
where $\text{Seed}_s \in \mathbb{R}^{K_s \times d}$ is a learnable seed embedding, and $\text{PE}(K_s)$ adds sinusoidal positional encodings. PMA allows the model to learn which features to group together, adapting to dataset-specific correlation structures.

\subsubsection{Special Tokens Injection:}

For each row at each scale, we prepend special tokens:

\begin{enumerate}
        \item $\text{CLS} \in \mathbb{R}^{N_{\text{cls}} \times d}$ (learnable, per-row summary)
    
        \item $\text{GLOBAL} \in \mathbb{R}^{N_{\text{global}} \times d}$ (learnable, long-range connectivity)
\end{enumerate}
    
The full sequence at scale $s$ becomes:
    
\begin{equation}
    X_s = [\text{CLS}, \text{GLOBAL}, \text{G}^{(s)}] \in \mathbb{R}^{B \times n \times (N_{\text{special}} + K_s) \times d}
\end{equation}    
where $N_{\text{special}} = N_{\text{cls}} + N_{\text{global}}$.

\subsubsection{Block-Sparse Attention Mask:}

As depicted in Figure \ref{fig:sparse}, for each scale, we construct a sparse attention mask $\mathbf{M}_s \in \mathbb{R}^{L_s \times L_s}$ where $L_s = N_{\text{special}} + K_s$. The mask follows these rules:
    
\begin{enumerate}
    \item \textbf{Fully Connected Special Tokens:} The first $N_{\text{special}}$ tokens (CLS and GLOBAL) are fully connected to all other tokens and to each other (Figure \ref{fig:sparse}.a):
    \begin{equation}
        \mathbf{M}_s[i, j] = 0 \quad \forall i \in [1, N_{\text{special}}] \text{ or } j \in [1, N_{\text{special}}]
    \end{equation}

    \item \textbf{Sliding Window Attention:} Feature tokens (indices $> N_{\text{special}}$) attend to neighbors within a window of radius $w=8$ (Figure \ref{fig:sparse}.b):
    \begin{equation}
        \mathbf{M}_s[i, j] = \begin{cases}
        0 & \text{if } |i - j| \leq w \text{ and } i, j > N_{\text{special}} \\
        -\infty & \text{otherwise}
    \end{cases}
        \end{equation}
    
    \item \textbf{Random Links (Optional):} For each feature token $i > N_{\text{special}}$, we randomly select $r$ additional tokens to attend to (Figure \ref{fig:sparse}.c):
    \begin{equation}
        \mathbf{M}_s[i, j_k] = 0 \quad \text{for } k \in [1, r], \, j_k \sim \text{Uniform}(\{N_{\text{special}}+1, \ldots, L_s\} \setminus \{i\})
    \end{equation}
    
    The final mask is (Figure \ref{fig:sparse}.d):
    \begin{equation}
        \mathbf{M}_s[i, i] = 0 \quad \forall i \in [1, L_s] \quad \text{(self-attention always allowed)}
    \end{equation} 
\end{enumerate}

\subsubsection{Transformer Encoder per Scale:}
    
For each scale $s \in \mathcal{S}$, we apply a dedicated Transformer encoder:

\begin{equation}
    \mathbf{Z}_s = \text{Encoder}_s(\mathbf{X}_s, \mathbf{M}_s) \in \mathbb{R}^{B \times n \times L_s \times d}
\end{equation}
    
where $\text{Encoder}_s$ consists of $N_{\text{blocks}}^{\text{row}} / |\mathcal{S}|$ stacked Transformer blocks with:
    
\begin{itemize}
    \item Multi-head self-attention: $\text{MHA}(\mathbf{Q}, \mathbf{K}, \mathbf{V}, \mathbf{M}_s)$ with sparse mask $\mathbf{M}_s$
    \item Rotary positional encoding (RoPE): Applied to queries and keys before attention~\cite{roformer}
    \item Feed-forward network: Two-layer MLP with GELU activation
    \item Pre-norm architecture: Layer normalization before each sub-layer
\end{itemize}
    
The multi-head attention with sparse masking is computed as:
    
\begin{align}
    \text{head}_i &= \text{Attention}(\mathbf{Q}_i, \mathbf{K}_i, \mathbf{V}_i, \mathbf{M}_s) \\
    &= \text{softmax}\left(\frac{\mathbf{Q}_i \mathbf{K}_i^\top}{\sqrt{d_k}} + \mathbf{M}_s\right) \mathbf{V}_i
\end{align}
    
where $\mathbf{M}_s$ contains $0$ for allowed positions and $-\infty$ for disallowed positions (additive masking).
    
After processing through $\text{Encoder}_s$, we extract the CLS token representations:
\begin{equation}
    \mathbf{H}_s = \mathbf{Z}_s[:, :, 1:N_{\text{cls}}, :] \in \mathbb{R}^{B \times n \times N_{\text{cls}} \times d}
\end{equation}
    
These represent the row-wise features at scale $s$. We then aggregate these representations across all scales by averaging:
\begin{equation}
    \mathbf{H}_{\text{agg}} = \frac{1}{|\mathcal{S}|} \sum_{s \in \mathcal{S}} \mathbf{H}_s \in \mathbb{R}^{B \times n \times N_{\text{cls}} \times d}
\end{equation}
    
This simple averaging strategy ensures that each scale contributes equally, balancing fine-grained and coarse-grained information. Next, the CLS tokens are flattened and normalized to produce the final row embeddings:
\begin{equation}
    \mathbf{H} = \text{LayerNorm}(\text{Flatten}(\mathbf{H}_{\text{agg}})) \in \mathbb{R}^{B \times n \times (N_{\text{cls}} \cdot d)}
\end{equation}
    
where $\text{Flatten}$ concatenates the $N_{\text{cls}}$ token embeddings.

Algorithm~\ref{alg:multi_scale_row} summarizes the complete multi-scale sparse row-wise interaction process.

\begin{algorithm}[h]
\caption{Multi-Scale Sparse Row-Wise Interaction ($\text{TF}_{\text{row}}^{\text{MS}}$)}
\label{alg:multi_scale_row}
\begin{algorithmic}[1]
\REQUIRE Embeddings $\mathbf{E} \in \mathbb{R}^{B \times n \times (m+C) \times d}$, valid features $d_{\text{valid}} \in \mathbb{R}^B$
\REQUIRE Scales $\mathcal{S} = \{s_1, s_2, \ldots, s_M\}$, window $w$, random links $r$
\ENSURE Row embeddings $\mathbf{H} \in \mathbb{R}^{n \times m \times (N_{\text{cls}} \cdot d)}$

\STATE Initialize learnable tokens $\mathbf{CLS} \in \mathbb{R}^{N_{\text{cls}} \times d}$, $\mathbf{GLOBAL} \in \mathbb{R}^{N_{\text{global}} \times d}$
\STATE $\mathbf{H}_{\text{all}} \gets []$ \COMMENT{Store CLS outputs from all scales}

\FOR{each scale $s \in \mathcal{S}$}
    \STATE $K_s \gets \lceil m / s \rceil$ \COMMENT{Number of groups at scale $s$}
    \STATE \textbf{// Feature Grouping}
    \STATE $\mathbf{G}^{(s)} \gets \text{PMA}(\mathbf{E}[:,:,C:,:], K_s)$     
    \STATE \textbf{// Construct Sequence}
    \STATE $\mathbf{X}_s \gets [\mathbf{CLS}, \mathbf{GLOBAL}, \mathbf{G}^{(s)}]$ \COMMENT{Shape: $(B, n, L_s, d)$ where $L_s = N_{\text{special}} + K_s$}
    
    \STATE \textbf{// Build Sparse Mask}
    \STATE $\mathbf{M}_s \gets \text{BuildBlockSparseMask}(L_s, N_{\text{special}}, w, r)$ 
    
    \STATE \textbf{// Process Through Transformer}
    \STATE $\mathbf{Z}_s \gets \text{Encoder}_s(\mathbf{X}_s, \mathbf{M}_s)$ \COMMENT{Transformer with RoPE and sparse attention}
    
    \STATE \textbf{// Extract CLS Tokens}
    \STATE $\mathbf{H}_s \gets \mathbf{Z}_s[:, :, 1:N_{\text{cls}}, :]$ 
    \STATE $\mathbf{H}_{\text{all}}.\text{append}(\mathbf{H}_s)$
\ENDFOR

\STATE \textbf{// Aggregate Across Scales}
\STATE $\mathbf{H}_{\text{agg}} \gets \frac{1}{|\mathcal{S}|} \sum_{s=1}^{|\mathcal{S}|} \mathbf{H}_{\text{all}}[s]$ 

\STATE \textbf{// Flatten and Normalize}
\STATE $\mathbf{H} \gets \text{LayerNorm}(\text{Flatten}(\mathbf{H}_{\text{agg}}))$ 

\RETURN $\mathbf{H}$
\end{algorithmic}
\end{algorithm}

\subsubsection{Computational Complexity}
For a given scale $s$ with $K_s = \lceil m / s \rceil$ grouped feature tokens, the per-layer computational complexity of the sparse attention mechanism is:

\begin{equation}
    O(B \cdot n \cdot L_s \cdot (w + N_{\text{global}} + r) \cdot d)
\end{equation}
where $B$ is the batch size, $n$ the number of samples per dataset, $m$ the number of features, $d$ the embedding dimension, and $w$, $N_{\text{global}}$, and $r$ denote the sliding-window size, number of global tokens, and number of random links, respectively.

For $M$ scales and a total of $N_{\text{blocks}}^{\text{row}}$ Transformer layers distributed evenly across scales, the overall complexity becomes:

\begin{equation}
O_{\text{total}} = \sum_{s \in \mathcal{S}} O(B \cdot n \cdot K_s \cdot (w + N_{\text{global}} + r) \cdot d \cdot \frac{N_{\text{blocks}}^{\text{row}}}{M})
\end{equation}

Since $\sum_{s \in \mathcal{S}} K_s \approx m \cdot \left(1 + \frac{1}{s_2} + \ldots + \frac{1}{s_M}\right)$ and typically $w, N_{\text{global}}, r \ll m$, this simplifies to:

\begin{equation}
O_{\text{total}} \approx O(B \cdot n \cdot m \cdot (w + N_{\text{global}} + r) \cdot d \cdot N_{\text{blocks}}^{\text{row}})
\end{equation}

compared to the dense attention cost of $O(B \cdot n \cdot m^2 \cdot d \cdot N_{\text{blocks}}^{\text{row}})$.

For typical hyperparameters ($m \in [10, 100]$, $w=8$, $N_{\text{global}}=4$, $r=2$), this results in a reduction from quadratic $O(m^2)$ to near-linear $O(m \cdot 14)$ complexity—achieving linear scaling while preserving both local and global feature dependencies.

\subsection{Cross-Component Memory with Perceiver Architecture}
\label{app:cross_component}
While the column-wise embedding and row-wise interaction components of tabular transformers independently model feature- and sample-level dependencies, richer contextual understanding can emerge if information is shared across these components. However, direct cross-component communication poses a major risk to the in-context learning (ICL) paradigm: naive attention between components can leak test-set information, violating the principle that predictions for test samples must depend solely on training examples and the test input itself.

To overcome this limitation, we introduce a Perceiver-style latent memory module \cite{perceiver} that enables safe, leak-free communication between architectural components. This latent memory acts as a shared representation space that can be written to by training samples and read from by both training and test samples, ensuring compliance with ICL constraints while promoting global knowledge sharing.

In standard transformer-based tabular architectures such as TabICL \cite{tabicl}, model components operate in a strictly sequential and isolated fashion:

\begin{enumerate}
    \item Column Embedding ($\text{TF}_{\text{col}}$): Encodes feature-wise statistics across samples to capture column-level distributions.
    \item Row Interaction ($\text{TF}_{\text{row}}$): Models dependencies across features within each sample.
    \item ICL Prediction ($\text{TF}_{\text{icl}}$): Performs in-context learning to infer test labels from training examples.
\end{enumerate}

This separation simplifies optimization and ensures ICL safety, but also introduces significant limitations:

\begin{itemize}
    \item No backward adaptation: Column embeddings cannot adjust based on row-level feature interactions.
    \item Limited contextual refinement: Row-level interactions lack access to global, dataset-level statistics beyond static column embeddings.
    \item Dataset isolation: Each dataset is processed independently, preventing cross-dataset generalization within a batch.
\end{itemize}

A fundamental ICL constraint is that test samples must not influence the model’s internal state in a way that affects training representations. Formally, letting

\begin{equation}
    \mathcal{D}_{\text{train}} = \{(\mathbf{x}_i, y_i)\}_{i=1}^{n_{\text{train}}}, \mathcal{D}_{\text{test}} = \{\mathbf{x}_j\}_{j=n_{\text{train}}+1}^{n}
\end{equation}
the prediction for a test sample $\mathbf{x}_j$ must satisfy:

\begin{equation}
\mathbb{P}(\hat{y}_j \mid \mathcal{D}_{\text{train}}, \mathcal{D}_{\text{test}}) = \mathbb{P}(\hat{y}_j \mid \mathcal{D}_{\text{train}}, \mathbf{x}_j)
\end{equation}

That is, the prediction depends only on the training set and the test features, never on other test representations or their labels.

Inspired by the Perceiver architecture~\cite{perceiver}, we introduce a learnable latent memory $\mathbf{L} \in \mathbb{R}^{P \times d}$ with $P$ memory slots. The key idea is:

\begin{enumerate}
    \item Write Phase (train-only): Memory attends to training representations to extract relevant global patterns.
    \item Read Phase (all samples): Both training and test samples attend to the memory to retrieve learned context, but cannot modify it.
\end{enumerate}

This asymmetry guarantees ICL safety, since only training data influence the memory’s contents. The memory serves as a compressed, permutation-invariant summary of the training context that enables consistent feature refinement across samples.

The memory module is incorporated inside the ICL transformer ($\text{TF}_{\text{icl}}$), refining the row embeddings before label injection and prediction.
Given row embeddings $\mathbf{H} \in \mathbb{R}^{B \times n \times d_h}$ (where $B$ is the batch size and $n$ the number of samples per dataset), the Perceiver memory transformation produces refined representations:

\begin{equation}
\mathbf{R} = \text{PerceiverMemory}(\mathbf{H}, n_{\text{train}}) \in \mathbb{R}^{B \times n \times d_H}
\end{equation}

with:

\begin{itemize}
    \item $d_H = N_{\text{cls}} \cdot d$ the hidden dimension after multi-head projection,

    \item $P$ the number of latent memory slots (a hyperparameter),
    
    \item and $n_{\text{train}}$ the number of labeled training examples.
    
\end{itemize}

The Perceiver memory consists of three key stages, each composed of multiple cross-attention layers with residual connections and feed-forward transformations.

\begin{enumerate}
    \item \textbf{Latent Memory Initialization:} We initialize a set of $P$ learnable latent vectors:
    \begin{equation}
    \mathbf{L}_0 \in \mathbb{H}^{P \times d_H}
    \end{equation}
    drawn from a truncated normal distribution $\mathcal{N}(0, 0.02^2)$. These latents act as a universal memory bank, shared across all datasets in the batch and reused across forward passes, providing a stable foundation for information aggregation.\\

    \item \textbf{Cross-Attention Block:} At the core of the memory is a cross-attention mechanism allowing one representation to attend to another.
    Given query set $\mathbf{Q}$ and key–value set $\mathbf{KV}$, we define:
    \begin{equation}
    \text{CrossAttn}(\mathbf{Q}, \mathbf{KV}) = \text{softmax}\left(\frac{\mathbf{Q} \mathbf{W}_Q (\mathbf{KV} \mathbf{W}_K)^\top}{\sqrt{d_k}}\right) (\mathbf{KV} \mathbf{W}_V)
    \end{equation}
    where $\mathbf{W}_Q, \mathbf{W}_K, \mathbf{W}_V \in \mathbb{R}^{d_R \times d_k}$ are projection matrices and $d_k = d_R / h$ is the per-head dimension.\\

    Each cross-attention block is followed by layer normalization, residual connections, and a feed-forward layer:
    \begin{align}
    \mathbf{Q}' &= \text{LayerNorm}(\mathbf{Q}) \\
    \mathbf{KV}' &= \text{LayerNorm}(\mathbf{KV}) \\
    \mathbf{Z} &= \mathbf{Q} + \text{MultiHeadCrossAttn}(\mathbf{Q}', \mathbf{KV}') \\
    \mathbf{Z}' &= \mathbf{Z} + \text{FFN}(\text{LayerNorm}(\mathbf{Z})).
    \end{align}
    This block structure ensures stable training and supports multi-head feature integration.\\

    \item \textbf{Write Phase: Memory Encoding:} In the write phase, the memory attends to \textit{training samples only} to extract and store relevant patterns. For each dataset $b$ in the batch:
    \begin{equation}
    \mathbf{H}_{\text{train}}^{(b)} = \mathbf{H}^{(b)}[:T_{\text{train}}, :] \in \mathbb{R}^{n_{\text{train}} \times d_H}
    \end{equation}
    and initialize the dataset-specific memory as $\mathbf{L}^{(b)}_0 = \mathbf{L}_0$.

    We then apply $N_{\text{write}}$ cross-attention blocks where memory latents query the training representations:
    
    \begin{equation}
    \mathbf{L}^{(b)}_{i+1} = \text{CrossAttnBlock}(\mathbf{Q} = \mathbf{L}^{(b)}_i, \mathbf{KV} = \mathbf{H}_{\text{train}}^{(b)}) \quad \text{for } i = 0, \ldots, N_{\text{write}} - 1
    \end{equation}
    
    The final encoded memory is:   
    \begin{equation}
    \mathbf{L}^{(b)} = \mathbf{L}^{(b)}_{N_{\text{write}}} \in \mathbb{R}^{P \times d_H}
    \end{equation}
    Importantly, $\mathbf{L}^{(b)}$ depends only on training representations, ensuring no test leakage.\\

    \item \textbf{Read Phase: Sample Refinement:} In the read phase, \textit{all samples} (training and test) attend to the memory to retrieve stored context. For dataset $b$:

    \begin{equation}
    \mathbf{H}^{(b)}_0 = \mathbf{H}^{(b)} \in \mathbb{R}^{n \times d_H}
    \end{equation}
    
    We apply $N_{\text{read}}$ cross-attention blocks where sample queries attend to the memory:
    
    \begin{equation}
    \mathbf{R}^{(b)}_{i+1} = \text{CrossAttnBlock}(\mathbf{Q} = \mathbf{R}^{(b)}_i, \mathbf{KV} = \mathbf{L}^{(b)}) \quad \text{for } i = 0, \ldots, N_{\text{read}} - 1
    \end{equation}
    
    The final refined embeddings are:
    
    \begin{equation}
    \mathbf{R}^{(b)} = \mathbf{R}^{(b)}_{N_{\text{read}}} \in \mathbb{R}^{n \times d_R}
    \end{equation}
    
\end{enumerate}

This asymmetric read–write design preserves the integrity of in-context learning:

\begin{itemize}
    \item Only training samples write to the memory.

    \item Both training and test samples read from it.

    \item The memory functions as a shared, compressed abstraction of the training data that can be safely leveraged for inference.
\end{itemize}

The complete ICL forward pass with Perceiver memory is described in Algorithm \ref{alg:icl_perceiver}:

\begin{algorithm}[h]
\caption{ICL with Perceiver Memory}
\label{alg:icl_perceiver}
\begin{algorithmic}[1]
\REQUIRE Row embeddings $\mathbf{H} \in \mathbb{R}^{B \times n \times d_H}$, training labels $\mathbf{y}_{\text{train}} \in \mathbb{R}^{B \times n_{\text{train}}}$
\ENSURE Predictions $\hat{\mathbf{y}} \in \mathbb{R}^{B \times (n - n_{\text{train}}) \times C}$ for $C$ classes

\STATE \textbf{// Perceiver Memory (optional)}
\IF{$P > 0$}
    \FOR{each dataset $b = 1$ to $B$}
        \STATE $\mathbf{H}_{\text{train}}^{(b)} \gets \mathbf{H}^{(b)}[:T_{\text{train}}, :]$ \COMMENT{Extract training samples}
        \STATE $\mathbf{L}^{(b)} \gets \mathbf{L}_0$ \COMMENT{Initialize memory}
        
        \STATE \textbf{// Write: Memory attends to training samples}
        \FOR{$i = 1$ to $N_{\text{write}}$}
            \STATE $\mathbf{L}^{(b)} \gets \text{CrossAttnBlock}(\mathbf{L}^{(b)}, \mathbf{H}_{\text{train}}^{(b)})$ 
        \ENDFOR
        
        \STATE \textbf{// Read: All samples attend to memory}
        \STATE $\mathbf{R}^{(b)} \gets \mathbf{H}^{(b)}$
        \FOR{$i = 1$ to $N_{\text{read}}$}
            \STATE $\mathbf{R}^{(b)} \gets \text{CrossAttnBlock}(\mathbf{R}^{(b)}, \mathbf{L}^{(b)})$ 
        \ENDFOR
    \ENDFOR
    \STATE $\mathbf{R} \gets \mathbf{R}$ \COMMENT{Use refined embeddings}
\ENDIF

\STATE \textbf{// Label Injection (training samples only)}
\STATE $\mathbf{R}[:, :n_{\text{train}}, :] \gets \mathbf{H}[:, :n_{\text{train}}, :] + \text{OneHot}(\mathbf{y}_{\text{train}}) \mathbf{W}_{\text{label}}$ 

\STATE \textbf{// ICL Transformer with Split Mask}
\STATE $\mathbf{H} \gets \text{TF}_{\text{icl}}(\mathbf{H}, \text{attn\_mask} = n_{\text{train}})$ \COMMENT{Prevent test-to-train leakage}

\STATE \textbf{// Prediction Head}
\STATE $\mathbf{R} \gets \text{LayerNorm}(\mathbf{R})$
\STATE $\text{logits} \gets \text{FFN}_{\text{decoder}}(\mathbf{R}[:, n_{\text{train}}:, :])$ \COMMENT{Predict test labels only}

\RETURN $\text{logits}$
\end{algorithmic}
\end{algorithm}


\subsection{Dataset-wise In-Context Learning}
\label{app:tficl}

After column-wise embedding, multi-scale sparse row-wise interaction, and optional cross-component memory refinement, each sample is represented by a fixed-dimensional row embedding:

\begin{equation}
   \mathbf{R} \in \mathbb{R}^{B \times n \times d_R} 
\end{equation}
 
where $B$ is the number of datasets in the batch, $n$ the total number of samples per dataset, and $d_R$ the embedding dimension.

The final component, dataset-wise in-context learning ($\text{TF}_{\text{icl}}$), leverages these embeddings to predict test labels by conditioning on labeled training examples—all within a single forward pass and without any gradient-based parameter updates..

Formally, for each dataset $b$ in the batch:
\begin{align}
    \mathcal{D}_{\text{train}}^{(b)} = \{(\mathbf{R}_i^{(b)}, y_i^{(b)})\}_{i=1}^{n_{\text{train}}}\\
    \mathcal{D}_{\text{test}}^{(b)} = \{\mathbf{R}_j^{(b)}\}_{j=n_{\text{train}}+1}^{n}
\end{align}

The objective is to predict test labels 
$\hat{y}_j^{(b)}$ for $j > n_{\text{train}}$ using in-context reasoning from training examples only.

\begin{equation}
\hat{\mathbf{y}}_{\text{test}} = \text{TF}_{\text{icl}}(\mathbf{R}, \mathbf{y}_{\text{train}})
\end{equation}

The ICL module consists of three main stages:

\begin{enumerate}
    \item \textbf{Label Encoding and Injection:} To ensure consistency across datasets with potentially different label spaces, training labels $\mathbf{y}_{\text{train}} \in \mathbb{R}^{B \times n_{\text{train}}}$ are first normalized to contiguous indices:
    
    \begin{equation}
    \tilde{y}_i = \text{argsort}(\text{unique}(\mathbf{y}_{\text{train}}))[\mathbf{y}_{\text{train}}[i]]
    \end{equation}

    mapping any label set $\{2,5,9\} \rightarrow \{0,1,2\}$.\\

    Normalized labels are embedded using one-hot encoding followed by a linear projection:
    \begin{equation}
    \mathbf{e}_y = \text{OneHot}(\tilde{y}, C_{\text{max}}) \cdot \mathbf{W}_y \in \mathbb{R}^{d_R}
    \end{equation}
    
    where $C_{\text{max}}$ is the maximum number of classes (e.g., $C_{\text{max}} = 10$), and $\mathbf{W}_y \in \mathbb{R}^{C_{\text{max}} \times d_R}$ is a learned projection matrix.
    
    Label embeddings are injected \textit{only} into training samples via additive combination:
    
    \begin{equation}
    \mathbf{R}[:, :n_{\text{train}}, :] \gets \mathbf{R}[:, :n_{\text{train}}, :] + \mathbf{e}_y(\mathbf{y}_{\text{train}})
    \end{equation}
    ensuring test samples remain unaffected and ICL constraints are preserved.\\

    \item \textbf{Split-Masked Transformer:}
    The augmented embeddings $R$ are processed by a split-masked Transformer, enforcing ICL-safe attention between training and test samples. The attention mask $\mathbf{M}_{\text{split}}$ is defined as:
    
    \begin{equation}
    \mathbf{M}_{\text{split}}[i, j] = \begin{cases}
    0 & \text{if } i \leq n_{\text{train}} \text{ and } j \leq n_{\text{train}} \quad \text{(train-to-train)} \\
    0 & \text{if } i > n_{\text{train}} \text{ and } j \leq n_{\text{train}} \quad \text{(test-to-train)} \\
    -\infty & \text{if } i \leq n_{\text{train}} \text{ and } j > n_{\text{train}} \quad \text{(train-to-test: blocked)} \\
    0 & \text{if } i > n_{\text{train}} \text{ and } j > n_{\text{train}} \quad \text{(test-to-test)}
    \end{cases}
    \end{equation}

No leakage from test to train samples.
    \begin{itemize}
        \item Training samples attend only to other training samples (learn from labeled context).
        \item Test samples attend to training samples and other test samples (contextual reasoning).
        \item No leakage from test to train samples.
    \end{itemize} 
    
    The Transformer applies $N_{\text{icl}}$ blocks of multi-head self-attention and feed-forward layers:
    
    \begin{align}
    \mathbf{H}^{(0)} &= \mathbf{R} \\
    \mathbf{H}^{(\ell+1)} &= \text{TransformerBlock}(\mathbf{H}^{(\ell)}, \mathbf{M}_{\text{split}}) \quad \text{for } \ell = 0, \ldots, N_{\text{icl}} - 1
    \end{align}
    with the final output normalized via:
    \begin{equation}
        \mathbf{H} = \text{LayerNorm}(\mathbf{H}^{(N_{\text{icl}})})
    \end{equation}

    \item \textbf{Prediction head:} Test sample representations $\mathbf{H}[:, n_{\text{train}}:, :]$ are passed through a two-layer MLP decoder:
    \begin{align}
    \mathbf{z} &= \text{GELU}(\mathbf{H}[:, n_{\text{train}}:, :] \mathbf{W}_1 + \mathbf{b}_1) \in \mathbb{R}^{B \times n_{\text{test}} \times 2d_R} \\
    \text{logits} &= \mathbf{z} \mathbf{W}_2 + \mathbf{b}_2 \in \mathbb{R}^{B \times n_{\text{test}} \times C_{\text{max}}}
    \end{align}
    
    Predictions are obtained via softmax with temperature $\tau$:
    \begin{equation}
    \hat{\mathbf{y}}_{\text{test}} = \text{softmax}(\text{logits}[:, :, :K] / \tau)
    \end{equation}
    where $K$ is the number of classes in the current dataset (inferred from training labels), and $\tau = 0.9$ by default.\\

    When the number of classes $K > C_{\text{max}}$ (e.g., $K > 10$), we employ a \textbf{hierarchical classification} strategy:
    Grouping: Partition 
    \begin{enumerate}
        \item Grouping: Partition $K$ classes into $G = \lceil K / C_{\text{max}} \rceil$ balanced groups.
        \item First-level prediction: Predict which group a test sample belongs to.
        \item Second-level prediction: For each group, train a classifier on the subset of classes within that group.
        \item Combination: Multiply group probability with intra-group probability to obtain final prediction.
    \end{enumerate}

    This hierarchical mechanism preserves the ICL paradigm while scaling to hundreds of classes.
\end{enumerate}

\begin{algorithm}[h]
\caption{Dataset-wise In-Context Learning}
\label{alg:icl}
\begin{algorithmic}[1]
\REQUIRE Row embeddings $\mathbf{R} \in \mathbb{R}^{B \times n \times d_R}$, training labels $\mathbf{y}_{\text{train}} \in \mathbb{R}^{B \times n_{\text{train}}}$
\ENSURE Predictions $\hat{\mathbf{y}}_{\text{test}} \in \mathbb{R}^{B \times (n - n_{\text{train}}) \times K}$

\STATE \textbf{// Optional: Perceiver Memory}
\IF{memory enabled}
    \STATE $\mathbf{R} \gets \text{PerceiverMemory}(\mathbf{R}, n_{\text{train}})$
\ENDIF

\STATE \textbf{// Label Encoding and Injection}
\STATE $\tilde{\mathbf{y}}_{\text{train}} \gets \text{NormalizeLabels}(\mathbf{y}_{\text{train}})$ \COMMENT{Map to $\{0, 1, \ldots, K-1\}$}
\STATE $\mathbf{e}_y \gets \text{OneHotLinear}(\tilde{\mathbf{y}}_{\text{train}})$ \COMMENT{Shape: $(B, n_{\text{train}}, d_R)$}
\STATE $\mathbf{R}[:, :n_{\text{train}}, :] \gets \mathbf{R}[:, :n_{\text{train}}, :] + \mathbf{e}_y$

\STATE \textbf{// Split-Masked Transformer}
\STATE $\mathbf{M}_{\text{split}} \gets \text{BuildSplitMask}(n, n_{\text{train}})$ 
\STATE $\mathbf{H} \gets \text{TF}_{\text{icl}}(\mathbf{R}, \mathbf{M}_{\text{split}})$ 
\STATE $\mathbf{H} \gets \text{LayerNorm}(\mathbf{H})$

\STATE \textbf{// Prediction Head}
\STATE $\mathbf{H}_{\text{test}} \gets \mathbf{H}[:, n_{\text{train}}:, :]$ \COMMENT{Extract test representations}
\STATE $\text{logits} \gets \text{MLP}_{\text{decoder}}(\mathbf{H}_{\text{test}})$ \COMMENT{Shape: $(B, T_{\text{test}}, C_{\text{max}})$}
\STATE $\text{logits} \gets \text{logits}[:, :, :K]$ \COMMENT{Select active classes}
\STATE $\hat{\mathbf{y}}_{\text{test}} \gets \text{softmax}(\text{logits} / \tau)$

\RETURN $\hat{\mathbf{y}}_{\text{test}}$
\end{algorithmic}
\end{algorithm}

During pretraining, the model is trained with cross-entropy loss on test samples:

\begin{equation}
\mathcal{L} = -\frac{1}{B \cdot n_{\text{test}}} \sum_{b=1}^{B} \sum_{j=n_{\text{train}}+1}^{n} \log p(y_j^{(b)} \mid \mathbf{R}^{(b)}, \mathbf{y}_{\text{train}}^{(b)})
\end{equation}

Critically, gradients flow through the entire architecture (column embedding, row interaction, memory, ICL transformer, decoder) in an end-to-end manner, enabling the model to learn representations optimized for in-context learning.

\newpage
\section{Experimental Evaluation}
We conduct a comprehensive evaluation of Orion-MSP. Below, we describe our experimental setup and present detailed results.

\subsection{Experimental Setting}

\subsubsection{Benchmark Suites and Datasets.} 

Our experimental evaluation spans three widely recognized benchmark suites: \textsc{TALENT}~\cite{talent} (181 automatically discovered classification datasets), \textsc{OpenML-CC18}~\cite{openmlcc18} (72 curated datasets), and \textsc{TabZilla}~\cite{tabzilla} (36 heterogeneous tasks). Together, these benchmarks enable a comprehensive assessment across diverse tabular learning scenarios. In addition, we perform domain-specific evaluations in high-impact application areas such as healthcare and finance to examine the real-world relevance of our method. All experiments strictly follow the official dataset splits provided by each benchmark to ensure reproducibility and fairness.

For consistency across model families, results are reported only on the intersection of datasets available to all evaluated models within each benchmark suite. This unified evaluation protocol ensures that observed performance differences arise from methodological advances rather than variations in dataset coverage. After filtering, our evaluation encompasses 154 of 181 datasets from TALENT, 63 of 72 from OpenML-CC18, and 27 of 36 from TabZilla. A small number of datasets were excluded due to out-of-memory (OOM) errors or CUDA-related issues, primarily affecting TabPFN-based architectures even on H200 GPUs.

Finally, we emphasize that models with higher mean ranks may not always achieve the highest absolute accuracy or F1-scores on every dataset. Rankings based on accuracy are computed per dataset and then averaged across all datasets, providing a normalized indicator of overall consistency rather than peak task-specific performance. In contrast, absolute metrics highlight maximum achievable performance on individual tasks. Comprehensive dataset statistics are presented in Appendix~\ref{app:data}.

\subsubsection{Models and Baselines.}
We compare our model with six state-of-the-art tabular foundation models: \textsc{TabPFN} \cite{tabpfn}, \textsc{TabICL} \cite{tabicl}, \textsc{OrionBiX}, \textsc{Mitra}, \textsc{ContextTab} \cite{contexttab}, and \textsc{TabDPT} \cite{tabdpt}. In addition, we include established \texttt{traditional baselines} using autogloun~\cite{erickson2020autogluon} such as \textsc{XGBoost, LightGBM, CatBoost,} and \textsc{Random Forest} as strong reference models for comparison.

\subsubsection{Hardware Configuration.} 
Experiments are executed on NVIDIA L40S GPUs, with H200 GPUs used for memory-intensive cases. This infrastructure ensures consistent execution across all experiments while handling the computational demands of large transformer-based models.

\subsubsection{Evaluation Metrics.} 
Our evaluation considers two complementary aspects:

\textbf{Performance.}
We measure predictive capability using standard classification metrics—Accuracy (ACC), AUC-ROC, and weighted F1-score (F1)—computed across the benchmark suites TALENT, OpenML-CC18, and TabZilla. These benchmarks encompass datasets with diverse characteristics, including varying sample sizes, feature dimensionalities, and class balance, allowing a comprehensive assessment of model generalization.
It is important to clarify how \textsc{mean rank} values are derived. Within each benchmark suite, models are ranked by accuracy on every dataset (lower rank = better performance), and these per-dataset ranks are averaged to obtain the overall mean rank. 
Thus, a lower mean rank indicates stronger and more consistent performance across datasets, rather than the highest score on any single task. While absolute metrics (\textsc{accuracy}, \textsc{F1}) reflect peak task-level performance, mean rank provides a normalized measure of cross-dataset generalization consistency.

\textbf{Scalability.}
We further analyze model robustness as dataset complexity increases by examining performance trends with respect to sample size, feature dimensionality, and class imbalance. This analysis uses the same benchmark datasets, aggregated along these axes to reveal systematic scalability behaviors and guide practical model selection.

\subsection{Results}

\begin{table}[hpbt]
  \footnotesize
  \centering
  \caption{Performance comparison across three benchmark suites—TALENT, OpenML-CC18, and TabZilla. Ranks denote the mean rank based on accuracy per benchmark suite (lower is better). Metrics: ACC = Accuracy, F1 = Weighted F1. The “All” column reports the aggregated rank across all suites. Formatting: \textbf{1st place}; \underline{2nd place}.}
  
  \vspace{0.1in}
  \label{tab:overall}
  \begin{tabular}{l c ccc ccc ccc}
    \toprule
    \multirow{2}{*}{Models} 
      & \multicolumn{1}{c}{All} 
      & \multicolumn{3}{c}{TALENT} 
      & \multicolumn{3}{c}{OpenML-CC18} 
      & \multicolumn{3}{c}{TabZilla} \\
    \cmidrule(lr){2-2} \cmidrule(lr){3-5} \cmidrule(lr){6-8} \cmidrule(lr){9-11}
     & Rank & Rank & ACC & F1 & Rank & ACC & F1 & Rank & ACC & F1 \\
    \midrule
    XGBoost    & 6.70 & 6.02 & 0.8403 & 0.8360 & 5.89 & 0.8558 & 0.8537 & 6.07 & 0.8612 & 0.8326 \\
    CatBoost  & 6.43 & 5.57 & 0.8336 & 0.8259 & 6.25 & 0.8588 & 0.8520 & 7.13 & 0.8579 & 0.8384 \\
    Random Forest        & 7.38 & 6.15 & 0.8285 & 0.8209 & 6.36 & 0.8547 & 0.8497 & 8.42 & 0.8358 & 0.8399 \\
    LightGBM  & 6.78 & 6.11 & 0.8331 & 0.8245 & 6.18 & 0.8581 & 0.8493 & 5.25 & 0.8618 & 0.8211 \\
    TabICL        & 4.96 & 4.09 & \underline{0.8471} & \underline{0.8379} & 4.69 & 0.8667 & 0.8623 & 5.89 & 0.8734 & 0.8698 \\
    OrionBiX & 5.37 & 4.59 & 0.8346 & 0.8260 & 4.98 & 0.8653 & 0.8596 & 4.89 & 0.8728 & 0.8628 \\
    \textbf{OrionMSP}     & 3.58 & \textbf{3.26} & 0.8461 & 0.8360 & \textbf{4.12} & \textbf{0.8722} & \textbf{0.8676} & \textbf{3.84} & \textbf{0.8821} & \textbf{0.8786} \\
    \underline{TabPFN}        & 4.61 & \underline{3.72} & \textbf{0.8514} & \textbf{0.8412} & 4.76 & \underline{0.8714} & \underline{0.8663} & 4.86 & 0.8752 & 0.8716 \\
    Mitra         & 11.77 & 10.38 & 0.3921 & 0.2868 & 10.52 & 0.3614 & 0.2522 & 11.21 & 0.3152 & 0.1830 \\
    ContextTab    & 9.70 & 9.84 & 0.5474 & 0.4596 & 6.28 & 0.8639 & 0.8581 & 7.13 & 0.8389 & 0.8334 \\
    TabDPT        & 5.42 & 5.19 & 0.8408 & 0.8318 & \underline{4.64} & 0.8672 & 0.8625 & \underline{3.94} & \underline{0.8814} & \underline{0.8775} \\

    \bottomrule
  \end{tabular}
\end{table}

Table~\ref{tab:overall} summarizes results across the TALENT, OpenML-CC18, and TabZilla benchmark suites, reporting mean rank, classification accuracy (ACC), and weighted F1-score (F1) for all evaluated models.

Our experiments confirm that classical machine learning methods remain strong baselines, achieving mean accuracies between 0.833 and 0.861 with aggregated ranks around 6.0. In contrast, pretrained tabular foundation models (TFMs) demonstrate superior generalization, even without task-specific fine-tuning. Notably, our model, \textsc{Orion-MSP}, achieves the best overall zero-shot rank of 3.58, with ACC/F1 scores of 0.8461/0.8360 on TALENT, 0.8722/0.8676 on OpenML-CC18, and 0.8821/0.8786 on TabZilla.

\textsc{TabPFN} follows closely, attaining an overall rank of 4.61 and scores of 0.8514/0.8412 on TALENT and up to 0.8752/0.8716 on TabZilla. \textsc{TabDPT} ranks 5.42, achieving 0.8408/0.8318 on TALENT and 0.8814/0.8775 on TabZilla. By contrast, Mitra (rank 11.77, ACC < 0.40) and ContextTab (rank 9.70) perform substantially worse, highlighting the advantages of hierarchical multi-scale processing and efficient attention in Orion-MSP.

Overall, \textsc{TabPFN} and \textsc{Orion-MSP} emerge as the strongest models, with ACC ranging from 0.85 to 0.88 and ranks between 3.26 and 4.61. \textsc{Orion-MSP} peaks on OpenML-CC18 (rank 4.12, ACC 0.8722) and TabZilla (rank 3.84, ACC 0.8821), while \textsc{TabPFN} leads on TALENT (ACC 0.8514) and maintains stable performance across all benchmark suites.

To further investigate the sources of Orion-MSP’s performance gains, we analyze results across key dataset characteristics. All analyses partition datasets based on inherent properties rather than performance outcomes.

\textbf{Dataset Size.} Table~\ref{tab:size_analysis} reports model performance aggregated by dataset size: Small ($<1K$ samples), Medium (1K-10K), and Large ($>10$). Performance trends reveal that Orion-MSP consistently performs well across small, medium, and large datasets. Classical ML models such as XGBoost excel on large datasets due to abundant training examples, achieving the highest ACC/F1 in the $>10K$ sample category. Orion-MSP, however, maintains competitive performance across all size categories, outperforming most baselines on small and medium datasets. This demonstrates the ability of multi-scale sparse attention to generalize effectively in low-data regimes while scaling gracefully to larger datasets. TabPFN also performs strongly, particularly on medium-sized datasets, but Orion-MSP’s consistent performance across size scales highlights the robustness of its hierarchical and sparse design.

\begin{table}[pt]
  \centering
  \scriptsize
  \caption{Performance variation by dataset size across all benchmark suites. Rank denotes the accuracy-based ranking per size category (lower is better). ACC = Accuracy; F1 = Weighted F1-score, averaged across datasets within each size category: Small ($<$1K samples), Medium (1K–10K), and Large ($>$10K).}
  \vspace{0.1in}
  \label{tab:size_analysis}
  \begin{tabular}{l ccc ccc ccc}
    \toprule
    \multirow{2}{*}{Models} & \multicolumn{3}{c}{Small (<1K)} & \multicolumn{3}{c}{Medium (1K-10K)} & \multicolumn{3}{c}{Large (>10K)} \\
    \cmidrule(lr){2-4} \cmidrule(lr){5-7} \cmidrule(lr){8-10}
     & Rank & ACC & F1 & Rank & ACC & F1 & Rank & ACC & F1 \\
    \midrule
    XGBoost & 7.70 & 0.8168 & 0.7964 & 6.88 & 0.8363 & 0.8314 & 5.41 & \textbf{0.8969} & \textbf{0.8920} \\
    CatBoost & 7.88 & 0.8124 & 0.7935 & 6.47 & 0.8340 & 0.8264 & 5.48 & 0.8797 & 0.8733 \\
    Random Forest & 8.55 & 0.7988 & 0.8187 & 7.16 & 0.8285 & 0.8221 & 7.30 & 0.8694 & 0.8628 \\
    LightGBM & 7.80 & 0.8143 & 0.7789 & 6.94 & 0.8314 & 0.8226 & 5.63 & 0.8827 & 0.8764 \\
    TabICL & 6.04 & 0.8301 & \textbf{0.8338} & 4.77 & 0.8486 & 0.8398 & 4.61 & 0.8802 & 0.8743 \\
    OrionBiX & 6.32 & \underline{0.8330} & 0.8150 & 5.48 & 0.8348 & 0.8260 & \underline{4.42} & 0.8729 & 0.8670 \\
    OrionMSP & \underline{5.93} & 0.8232 & 0.8194 & \textbf{3.70} & \underline{0.8494} & \underline{0.8402} & \textbf{3.04} & \underline{0.8843} & \underline{0.8768} \\
    TabPFN & 6.50 & 0.8325 & 0.8131 & \underline{3.81} &  \textbf{0.8557} & \textbf{0.8462} & 5.73 & 0.8783 & 0.8713 \\
    Mitra & 13.88 & 0.4334 & 0.3236 & 11.59 & 0.3600 & 0.2553 & 11.11 & 0.3837 & 0.2754 \\
    ContextTab & 9.60 & 0.7578 & 0.7363 & 9.52 & 0.6210 & 0.5566 & 10.22 & 0.6388 & 0.5638 \\
    TabDPT & \textbf{5.48} & \textbf{0.8333} & \underline{0.8271} & 5.40 & 0.8424 & 0.8339 & 5.26 & 0.8831 & 0.8765 \\
    \bottomrule
  \end{tabular}
\end{table}

\textbf{Feature Dimensionality.} Table~\ref{tab:width_analysis} presents performance trends across narrow ($<10$ features), medium (10 - 100) and wide ( $>100$) datasets. When evaluating dataset width, Orion-MSP shows the highest accuracy on narrow datasets (<10 features) and strong performance on medium and wide datasets (10–100 and >100 features). This suggests that sparse multi-scale attention enables effective learning even in high-dimensional feature spaces, where dense models such as TabICL exhibit diminished scalability to high-dimensional feature spaces.
\begin{table}[pt]
  \centering
  \scriptsize
  \caption{Performance variation by feature dimensionality (dataset width) across all benchmark suites. Rank denotes the accuracy-based ranking averaged within each width category (lower is better). ACC = Accuracy; F1 = Weighted F1-score. Values are on a 0–1 scale (higher is better). Formatting: \textbf{1st place} ; \underline{2nd place} within each group.}
  \vspace{0.1in}
  \label{tab:width_analysis}
  \begin{tabular}{l ccc ccc ccc}
    \toprule
    \multirow{2}{*}{Models} & \multicolumn{3}{c}{Narrow (<10)} & \multicolumn{3}{c}{Medium (10-100)} & \multicolumn{3}{c}{Wide (>100)} \\
    \cmidrule(lr){2-4} \cmidrule(lr){5-7} \cmidrule(lr){8-10}
     & Rank & ACC & F1 & Rank & ACC & F1 & Rank & ACC & F1 \\
    \midrule
    XGBoost & 6.77 & 0.8222 & 0.8159 & 6.90 & 0.8482 & 0.8410 & \textbf{4.79} & 0.9140 & 0.9039 \\
    CatBoost & 5.63 & 0.8145 & 0.8067 & 6.88 & 0.8441 & 0.8344 & \underline{5.50} & \textbf{0.9157} & \underline{0.9084} \\
    Random Forest & 7.15 & 0.8005 & 0.7044 & 7.44 & 0.8410 & 0.8235 & 7.52 & 0.9034 & 0.8936 \\
    LightGBM & 6.15 & 0.8128 & 0.7907 & 6.92 & 0.8458 & 0.8326 & 7.47 & 0.8999 & 0.8908 \\
    \underline{TabICL} & 5.14 & 0.8208 & 0.8119 & 4.61 & \underline{0.8627} & \underline{0.8549} & 6.46 & 0.9101 & 0.8936 \\
    OrionBiX & \underline{4.64} & 0.8112 & 0.8043 & 5.46 & 0.8510 & 0.8417 & 6.73 & 0.8859 & 0.8849 \\
    \underline{OrionMSP} & \textbf{3.76} & \textbf{0.8394} & \textbf{0.8314} & \underline{4.09} & 0.8572 & 0.8478 & 5.69 & 0.8860 & 0.8837 \\
    \underline{TabPFN} & 5.30  & 0.8187 & 0.8092 & \textbf{4.07} & \textbf{0.8676} & \textbf{0.8589} & 6.141 & \underline{0.9129} & \textbf{0.9111} \\
    Mitra & 11.25 & 0.3737 & 0.2683 & 11.84 & 0.3886 & 0.2781 & 13.03 & 0.2521 & 0.1497 \\
    ContextTab & 9.52 & 0.6391 & 0.5719 & 9.59 & 0.6480 & 0.5843 & 10.97 & 0.6017 & 0.5651 \\
    \underline{TabDPT} & 4.66 & \underline{0.8262} & \underline{0.8189} & 5.45 & 0.8566 & 0.8483 & 7.23 & 0.8845 & 0.8820 \\
    \bottomrule
  \end{tabular}
\end{table}

\textbf{Based on Class Imbalance.} Partitioning datasets based on class balance reveals that Orion-MSP achieves its strongest gains on imbalanced datasets. The model ranks second in this category, achieving ACC = 0.8840 and F1 = 0.8731. This highlights that multi-scale sparse attention amplifies signals from underrepresented classes while avoiding overfitting to dominant classes. On balanced datasets, performance gains are smaller, suggesting that the architectural complexity of Orion-MSP is most advantageous when datasets exhibit skewed distributions. In comparison, TabPFN maintains strong performance on both balanced and imbalanced datasets, but Orion-MSP’s design more effectively addresses minority-class patterns due to hierarchical attention and cross-scale reasoning.

\begin{table}[pt]
  \centering
  \scriptsize
  \caption{Performance variation by class imbalance across all benchmark suites. ACC = Accuracy; F1 = Weighted F1-score, averaged within each imbalance category. Rank denotes the mean rank within each category (lower is better). Formatting: \textbf{1st place} ; \underline{2nd place} within each group.}
  \vspace{0.1in}
  \label{tab:imbalance_analysis}
  \begin{tabular}{l cc cc cc}
    \toprule
    \multirow{2}{*}{Models} & \multicolumn{3}{c}{Balanced ($\ge 0.6$)} & \multicolumn{3}{c}{Imbalanced (<0.6)} \\
    \cmidrule(lr){2-4} \cmidrule(lr){5-7}
     & Rank & ACC & F1 & Rank & ACC & F1 \\
    \midrule
    XGBoost & 7.00 & 0.8175 & 0.8110 & 6.23 & \textbf{\underline{0.8859}} & \textbf{\underline{0.8785}} \\
    CatBoost & 7.15 & 0.8076 & 0.8020 & 5.65 & 0.8785 & 0.8665 \\
    Random Forest & 7.92 & 0.7983 & 0.7955 & 6.77 & 0.8741 & 0.8646 \\
    LightGBM & 7.32 & 0.8071 & 0.7977 & 6.19 & 0.8775 & 0.8633 \\
    TabICL & 4.72 & \underline{0.8279} & \underline{0.8233} & 5.08 & 0.8806 & 0.8698 \\
    OrionBiX & 5.65 & 0.8096 & 0.8040 & \underline{5.04} & 0.8787 & 0.8683 \\
    \underline{OrionMSP} & \underline{4.22} & 0.8265 & 0.8202 & \textbf{\underline{3.38}} & \underline{0.8840} & \underline{0.8731} \\
    \textbf{\underline{TabPFN}} & \textbf{\underline{3.85}} & \textbf{\underline{0.8367}} & \textbf{\underline{0.8309}} & 5.37 & 0.8808 & 0.8697 \\
    Mitra & 12.26 & 0.2763 & 0.1540 & 11.24 & 0.4794 & 0.3858 \\
    ContextTab & 9.66 & 0.5079 & 0.4487 & 9.72 & 0.7850 & 0.7192 \\
    TabDPT & 5.16 & 0.8233 & 0.8189 & 5.65 & 0.8798 & 0.8690 \\
    \bottomrule
  \end{tabular}
\end{table}

\textbf{Domain-specific Analysis.} Domain-wise evaluation provides deeper insight into Orion-MSP’s strengths (Table~\ref{tab:domains_mf}):

\begin{itemize}
    \item Medical datasets: Orion-MSP achieves the highest ACC = 0.8045 and F1 = 0.7916, ranking second overall behind Orion-BiX. These datasets often involve hierarchical biological structures and complex interdependencies among features, which align naturally with Orion-MSP’s multi-scale representation. Fine-grained scales capture local dependencies, while coarser scales aggregate contextual information, leading to improved predictive accuracy.

    \item Finance datasets: Orion-MSP ranks first in mean rank (4.60), achieving ACC = 0.8158 and F1 = 0.8047. Financial datasets frequently involve layered dependencies between assets, instruments, and market indicators. Orion-MSP’s cross-component memory allows information to propagate across scales, capturing global dependencies that standard dense transformers or classical ML models fail to exploit.

\end{itemize}

Overall, domain-specific results highlight that Orion-MSP excels in high-dimensional, context-rich datasets, where hierarchical patterns and feature correlations are prevalent.

\begin{table}[hbpt]
  \centering
  \scriptsize
  \caption{Domain-specific performance for Medical and Finance datasets from the benchmark suites. Rank denotes the mean rank within each domain (lower is better). ACC = Accuracy; F1 = Weighted F1-score (0–1 scale, higher is better). Formatting: \textbf{1st place}; \underline{2nd place} within each group.}
  \vspace{0.1in}
  \label{tab:domains_mf}
  \begin{tabular}{l ccc c ccc}
    \toprule
    \multirow{2}{*}{Models} & \multicolumn{3}{c}{Medical} & \multicolumn{3}{c}{Finance} \\
    \cmidrule(lr){2-4} \cmidrule(lr){5-7}
     & Rank & ACC & F1 & Rank & ACC & F1 \\
    \midrule
    XGBoost & 6.32 & 0.7834 & 0.7669 & 6.62 & 0.7958 & 0.7885 \\
    RandomForest & 6.38 & 0.7779 & 0.7752 & 7.32 & 0.8052 & 0.8001 \\
    CatBoost & 6.36 & 0.7784 & 0.7594 & 5.82 &  0.8117 & 0.8015 \\
    LightGBM & 5.32 & 0.7949 & 0.7614 & 6.17 & 0.8095 & 0.7974 \\
    TabICL & 5.54 & 0.7819 & 0.7696 & 6.60 & 0.8125 & 0.7942 \\
    \textbf{\underline{OrionBiX}} & \textbf{\underline{4.10}} & 0.7893 & 0.7759 & \underline{5.39} & \textbf{\underline{0.8206}} & \textbf{\underline{0.8125}} \\
    \underline{OrionMSP} & \underline{4.50} & \textbf{\underline{0.8045}} & \textbf{\underline{0.7916}} & \textbf{\underline{4.60}} & \underline{0.8158} & \underline{0.8047} \\
    TabPFN & 5.04 & \underline{0.7984} & \underline{0.7857} & 7.17 & 0.8094 & 0.7919 \\
    Mitra & 10.77 & 0.3935 & 0.2863 & 13.67 & 0.5340 & 0.4250 \\
    ContextTab & 8.66 & 0.6681 & 0.6129 & 11.25 & 0.7430 & 0.6834 \\
    TabDPT & 6.86 & 0.7764 & 0.7641 & 8.00 & 0.8080 & 0.7960 \\
    \bottomrule
  \end{tabular}
\end{table}

\subsubsection*{Deep Analysis and Interpretation}

A detailed examination by dataset characteristics demonstrates why Orion-MSP’s design is most effective under certain conditions:

\begin{itemize}
    \item Class imbalance: Multi-scale sparse attention amplifies underrepresented patterns without overfitting to majority classes. Minority-class recognition improves substantially on datasets where the minority class constitutes less than 30\% of the data. Balanced datasets show smaller gains, indicating that the hierarchical complexity is most beneficial in skewed settings.

    \item Hierarchical structure and cross-component memory: In domains such as healthcare and finance, datasets involve natural hierarchies and complex inter-feature relationships. Orion-MSP’s multi-scale design allows it to reason at both fine-grained and coarse-grained levels. Sparse attention reduces computational cost and provides implicit regularization, mitigating overfitting in high-dimensional or correlated-feature settings. Cross-component memory further enables information exchange across scales without violating ICL safety, enhancing performance on context-dependent tasks.

    \item Computational efficiency: Linear attention complexity with respect to feature number and attention window size allows Orion-MSP to scale to high-dimensional tables. Memory usage grows proportionally with input dimensions, making the model practical for large real-world datasets, unlike dense attention alternatives with quadratic scaling.

\end{itemize}

In short, fine-grained scales capture subtle minority-class patterns, while coarser scales aggregate global context, yielding a balanced representations of local and global dependencies. Sparse attention improves efficiency and regularization, reducing overfitting in high-dimensional or correlated-feature settings. The Perceiver memory enhances the model’s capacity to store and retrieve non-local patterns, enabling cross-scale reasoning—particularly valuable in context-dependent domains. However, the added architectural complexity offers limited benefit for simpler, low-dimensional datasets, suggesting future directions in adaptive designs with data-driven scale selection and dynamic sparsity control.

\section{Conclusion}
In this work, we introduced Orion-MSP, a novel tabular in-context learning model that leverages multi-scale sparse attention and cross-component memory to capture both fine-grained and coarse-grained dependencies in tabular data. Through extensive experiments across diverse benchmark suites—including TALENT, OpenML-CC18, and TabZilla—as well as domain-specific datasets in healthcare and finance, we demonstrated that Orion-MSP consistently achieves state-of-the-art zero-shot performance, particularly on imbalanced, high-dimensional, and context-rich datasets.

Our detailed analyses highlight that the hierarchical design, sparse attention, and cross-component memory collectively contribute to robust generalization, efficient computation, and improved representation of complex interdependencies. These architectural choices enable Orion-MSP to outperform existing tabular foundation models in challenging real-world scenarios while maintaining practical scalability.

Nonetheless, we observe that the benefits of multi-scale sparse attention are less pronounced on simple, low-dimensional datasets, where the additional architectural complexity may not be fully leveraged. This limitation motivates future work on adaptive scale selection and data-aware sparsity scheduling, allowing model complexity to adjust dynamically to dataset characteristics. Such extensions could further enhance both efficiency and generality, enabling Orion-MSP to provide strong performance across the full spectrum of tabular learning tasks.

In summary, Orion-MSP represents a promising step toward scalable, adaptive, and context-aware tabular in-context learning, with significant potential for real-world applications and future improvements in dynamic model adaptation.

\bibliographystyle{plain}
\bibliography{refrences}

\appendix
\section{Pretraining and Implementation Details}
\label{app:pretraining}

\subsection{Pretraining Data Generation}
\label{app:pretraining_data}

Following the pretraining paradigm of TabICL~\cite{tabicl}, we train our model on synthetically generated datasets to learn generalizable representations for in-context learning on tabular data. Unlike natural language or vision domains where large-scale real data is available, tabular datasets exhibit extreme heterogeneity in schemas, distributions, and task objectives. Synthetic data generation via structural causal models (SCMs) enables us to control dataset diversity while ensuring coverage of diverse statistical patterns.

\subsubsection{Structural Causal Model (SCM) Prior}

We generate synthetic datasets using SCM-based priors \cite{scm,muller2022pfn}, where features are related through nonlinear causal relationships. For a dataset with $m$ features, we define a directed acyclic graph (DAG) $\mathcal{G} = (\mathcal{V}, \mathcal{E})$ where each node $v \in \mathcal{V}$ represents a feature, and edges $\mathcal{E}$ encode causal dependencies.

Each feature $X_j$ is computed as:

\begin{equation}
X_j = f_j(\text{Pa}(X_j), \epsilon_j)
\end{equation}

where $\text{Pa}(X_j)$ are the parent features of $X_j$ in $\mathcal{G}$, $f_j$ is a nonlinear activation function, and $\epsilon_j \sim \mathcal{N}(0, \sigma_j^2)$ is Gaussian noise.

\paragraph{Activation Function Diversity}

To ensure broad coverage of feature transformations observed in real-world tabular data, we used the following activation functions:

\begin{itemize}
    \item \textbf{Identity}, 
    \item $\mathbf{tanh}$, 
    \item \textbf{LeakyReLU}, 
    \item \textbf{ELU}
    \item \textbf{Standard nonlinearities}: ReLU, ReLU6~\cite{relu6}, SELU~\cite{selu}, SiLU (Swish)~\cite{silu}, Softplus
    \item \textbf{Bounded functions}: $\text{Hardtanh}(x) = \max(-1, \min(1, x))$, Signum function $\text{sgn}(x)$
    \item \textbf{Periodic functions}: $\sin(x)$ (captures cyclic patterns, e.g., time-of-day)
    \item \textbf{Radial basis function}: $\text{RBF}(x) = \exp(-x^2)$ (models local interactions)
    \item \textbf{Exponential growth/decay}: $\exp(x)$ (models compounding effects, e.g., financial data)
    \item \textbf{Power functions}: $f(x) = \sqrt{|x|}$, $f(x) = x^2$, $f(x) = |x|$ (models scaling relationships)
    \item \textbf{Indicator function}: $f(x) = \mathbb{1}_{|x| \leq 1}$ (models threshold effects)
    \item \textbf{Random Fourier features}: $f(x) = \boldsymbol{\phi}(x)^\top \mathbf{z}$ where $\mathbf{z} \sim \mathcal{N}(\mathbf{0}, \mathbf{I})$ and the feature map $\boldsymbol{\phi}: \mathbb{R} \to \mathbb{R}^N$ is defined as:
    \begin{equation}
    \phi_i(x) := \frac{w_i}{\|\mathbf{w}\|_2} \cdot \sin(a_i x + b_i), \quad i \in \{1, \ldots, N\}
    \end{equation}
    with $N = 256$, $b_i \sim \mathcal{U}[0, 2\pi]$, $a_i \sim \mathcal{U}[0, N]$, and $w_i := a_i^{-\exp(u)}$ where $u \sim \mathcal{U}[0.7, 3.0]$. This random function approximates complex, non-parametric relationships~\cite{rahimi2007random}.
\end{itemize}

For each feature $X_j$, an activation function $f_j$ is sampled uniformly from this extended set, ensuring diverse nonlinear transformations across the dataset.

\subsubsection{Tree-Based SCM Prior}

To complement the continuous SCM prior, we also generate datasets using a tree-based SCM prior \cite{grinsztajn2022tree}, where the causal mechanism $f_j$ is a decision tree or random forest. This prior is particularly important for modeling categorical interactions and hierarchical decision boundaries commonly observed in real-world tabular data (e.g., credit scoring, medical diagnosis).

For each feature $X_j$, we construct a random decision tree $\mathcal{T}_j$ with:
\begin{itemize}
    \item \textbf{Splitting criteria}: Random thresholds on parent features $\text{Pa}(X_j)$
    \item \textbf{Leaf values}: Sampled from a Gaussian or uniform distribution
    \item \textbf{Tree depth}: Sampled uniformly from $\{1, 2, 3, 4\}$ to vary complexity
\end{itemize}

The tree-based prior generates datasets with \textit{piecewise-constant} relationships, contrasting with the smooth transformations of the MLP-based SCM prior.

\subsection{Pretraining Details}
We employ a three-stage curriculum learning strategy that progressively increases dataset size (number of samples per dataset) and refines different architectural components. 

\begin{enumerate}
    \item \textbf{Stage 1} (25K steps, 2,048 datasets) trains all components end-to-end with $N_B = 8$ micro-batches for gradient accumulation, where each dataset contains a fixed size of 1,024 samples. This stage establishes foundational representations across column embedding, multi-scale sparse row interaction, Perceiver memory, and ICL prediction. 

    \item \textbf{Stage 2} (2K steps, 512 datasets) reduces micro-batch size to $N_B = 1$ and samples dataset sizes from a log-uniform distribution $\mathcal{U}_{\log}[1024, 40000]$, exposing the model to variable context lengths while maintaining architectural diversity. Within each micro-batch, all datasets share the same sample count, but this count varies across micro-batches. 

    \item \textbf{Stage 3} (50 steps, 512 datasets) focuses exclusively on long-context ICL by freezing all components except $\text{TF}_{\text{icl}}$ and sampling dataset sizes uniformly from $\mathcal{U}[40000, 60000]$, ensuring robust in-context learning on large datasets. 
\end{enumerate}

\begin{figure}[htbp]
  \centering
  \begin{subfigure}[b]{0.31\textwidth}
    \centering
    \includegraphics[width=\textwidth]{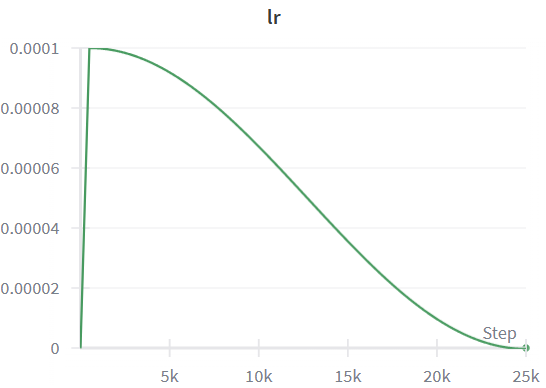}
    \caption{Cosine decay for stage 1}
    \label{fig:lr1}
  \end{subfigure}
  \hfill
  \begin{subfigure}[b]{0.31\textwidth}
    \centering
    \includegraphics[width=\textwidth]{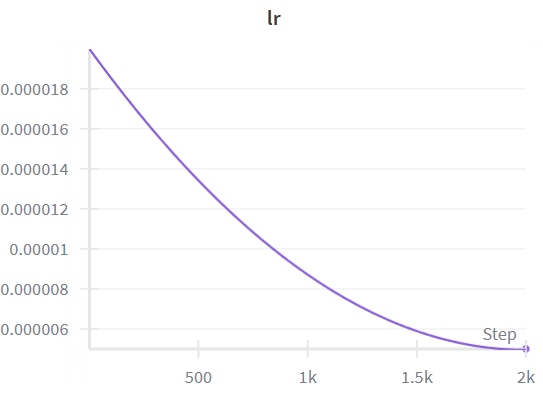}
    \caption{Polynomial decay for stage 2}
    \label{fig:lr2}
  \end{subfigure}
  \begin{subfigure}[b]{0.31\textwidth}
    \centering
    \includegraphics[width=\textwidth]{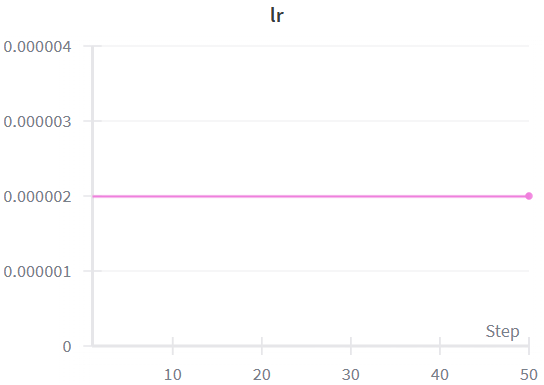}
    \caption{Constant lr for stage 3}
    \label{fig:lr3}
  \end{subfigure}

  \caption{Learning rate schedules for pretraining stages.}
  \label{fig:lr}
\end{figure}

Across all stages, we use the Adam optimizer~\cite{kingma2014adam} with gradient norm clipping at 1.0 and a learning rate schedule shown in Figure~\ref{fig:lr}. This curriculum, progressing from small, uniform datasets to large, variable datasets with selective fine-tuning, enables the model to generalize effectively across diverse dataset scales while preventing overfitting to specific sample counts.

\subsection{Implementation Details}

This section provides comprehensive hyperparameters and training configurations for all architectural components of ORION-MSP. Our implementation is built on PyTorch and trained on NVIDIA H200 GPUs.

\subsubsection{Column-wise Embedding (\texorpdfstring{$\text{TF}_{\text{col}}$}{TFcol})}

\paragraph{Hyperparameters}
\begin{itemize}
    \item Embedding dimension: $d = 128$
    \item Number of inducing points: $k = 128$
    \item Attention heads: $h = 4$ (head dimension $d_k = 32$)
    \item ISAB blocks: $L = 3$
    \item Feedforward dimension: $d_{\text{ff}} = 256$
    \item Dropout rate: $p = 0.0$
    \item Activation: GELU
    \item Layer norm: Pre-norm
\end{itemize}

\paragraph{Initialization and Training Considerations}
\begin{itemize}
    \item Inducing points initialized from $\mathcal{N}(0, 0.02^2)$
    \item Linear and attention weights: Xavier/Glorot uniform
    \item Column dropout: $p_{\text{col}} = 0.1$
    \item Gradient clipping: max-norm = 1.0
\end{itemize}

\subsubsection{Multi-Scale Sparse Row Interaction}

\paragraph{Hyperparameters}
\begin{itemize}
    \item Embedding dimension: $d = 128$
    \item Transformer blocks: $N_{\text{blocks}}^{\text{row}} = 6$ (2 per scale)
    \item Scales: $\mathcal{S} = \{1, 4, 16\}$
    \item Attention heads: $h = 8$
    \item Window size: $w = 8$, Global tokens: $N_{\text{global}} = 4$, Random links: $r = 2$
    \item Dropout: $p = 0.0$
    \item Positional encoding: RoPE ($\theta = 100000$)
\end{itemize}

\paragraph{Training Details}
\begin{itemize}
    \item Sparse attention via PyTorch SDPA
    \item Cosine learning rate schedule with 5\% warmup
    \item Mixed precision: FP16 (forward), FP32 (softmax)
\end{itemize}

\subsubsection{Cross-Component Perceiver Memory}

\paragraph{Hyperparameters}
\begin{itemize}
    \item Memory slots: $P = 32$
    \item Write layers: $N_{\text{write}} = 2$, Read layers: $N_{\text{read}} = 2$
    \item Attention heads: $h = 4$
    \item Feedforward dimension: $d_{\text{ff}} = 2d_R$
    \item Dropout: $p = 0.0$
    \item Activation: GELU, Layer norm: Pre-norm
\end{itemize}

\paragraph{Training Considerations}
\begin{itemize}
    \item Random initialization of memory $\mathbf{L}_0 \sim \mathcal{N}(0, 0.02^2)$
    \item Gradient clipping (max-norm = 1.0)
    \item Memory disabled ($P = 0$) for ablation
\end{itemize}

\subsubsection{Dataset-wise In-Context Learning}

\paragraph{Hyperparameters}
\begin{itemize}
    \item Transformer blocks: $N_{\text{icl}} = 12$
    \item Embedding dimension: $d_R = 512$
    \item Feedforward dimension: $d_{\text{ff}} = 1024$
    \item Max classes: $C_{\text{max}} = 10$
    \item Temperature: $\tau = 0.9$
    \item Dropout: $p = 0.0$
    \item Activation: GELU
    \item Layer norm: Pre-norm
\end{itemize}

\paragraph{Initialization}
\begin{itemize}
    \item Label encoder, Transformer, and decoder weights: Xavier/Glorot uniform
    \item Layer norm: $\gamma = 1$, $\beta = 0$
\end{itemize}

\section{Further Experiments}
\label{app:results}

\begin{figure}[htbp]
  \centering
  \begin{subfigure}[b]{0.75\textwidth}
    \centering
    \includegraphics[width=\textwidth]{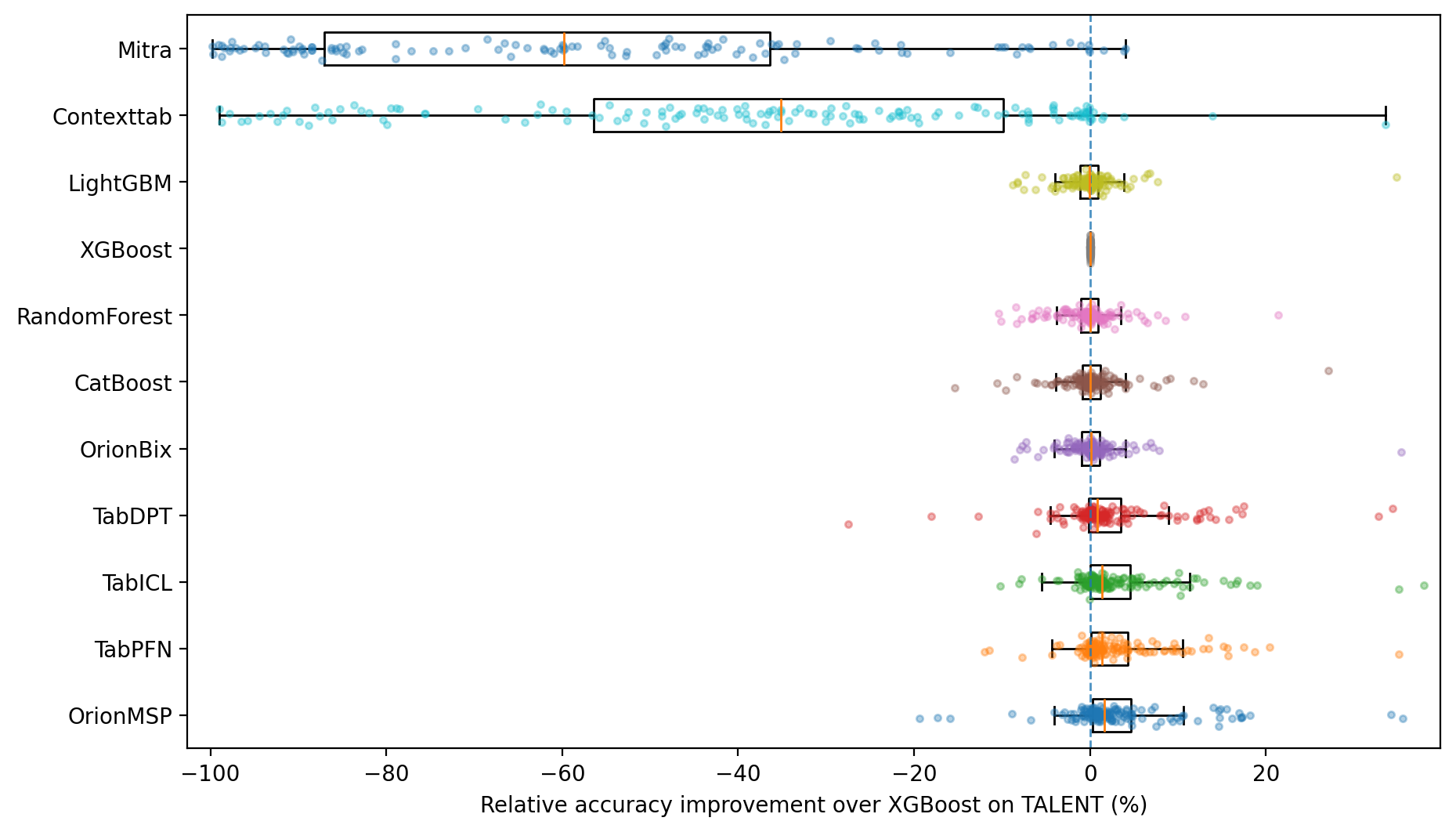}
    \caption{Relative accuracy improvement over XGBoost on TALENT Benchmark}
    \label{fig:imp_talent}
  \end{subfigure}
  \hfill
  \begin{subfigure}[b]{0.75\textwidth}
    \centering
    \includegraphics[width=\textwidth]{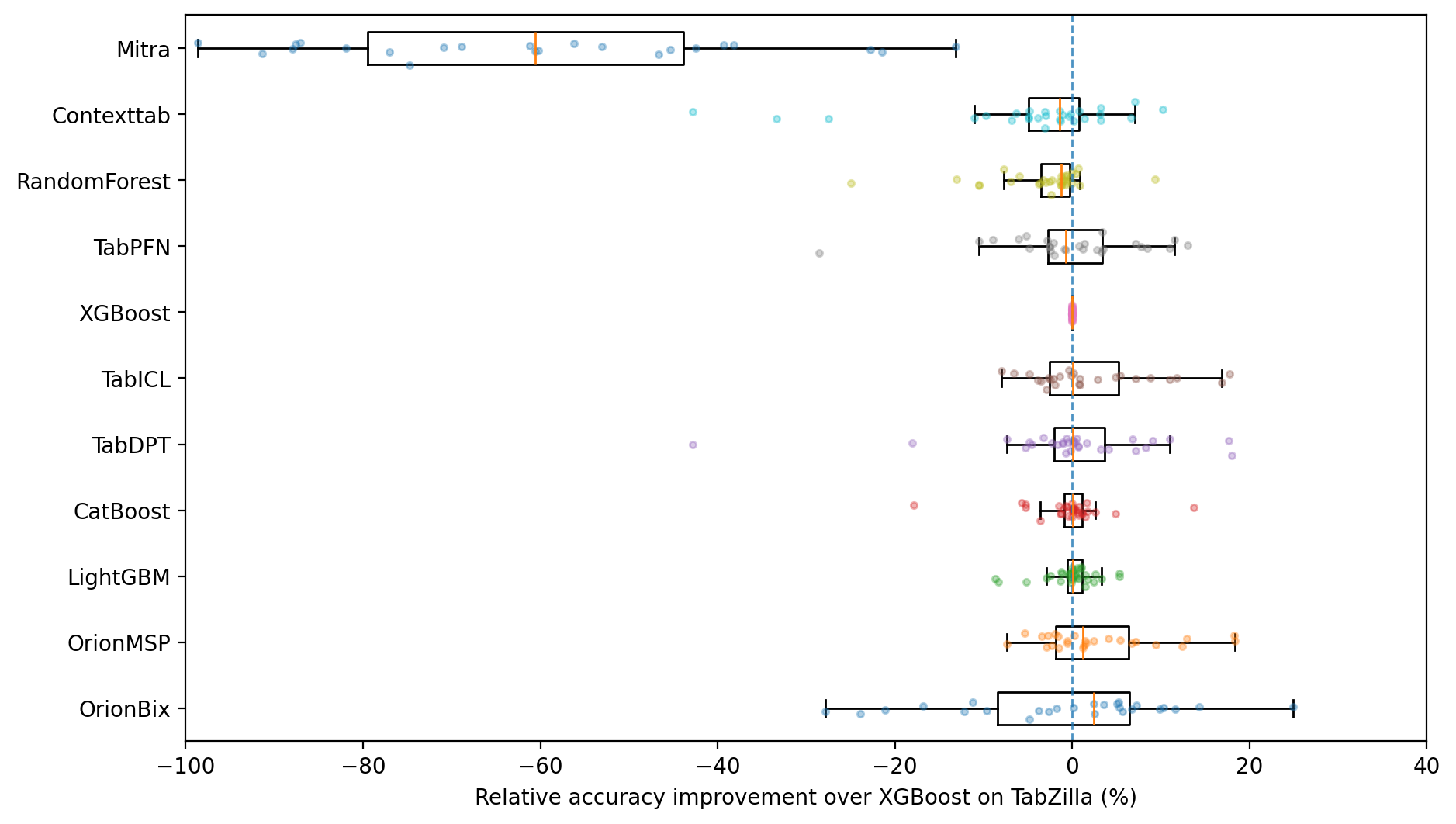}
    \caption{Relative accuracy improvement over XGBoost on TabZilla Benchmark}
    \label{fig:imp_tabzilla}
  \end{subfigure}
   \hfill
  \begin{subfigure}[b]{0.75\textwidth}
    \centering
    \includegraphics[width=\textwidth]{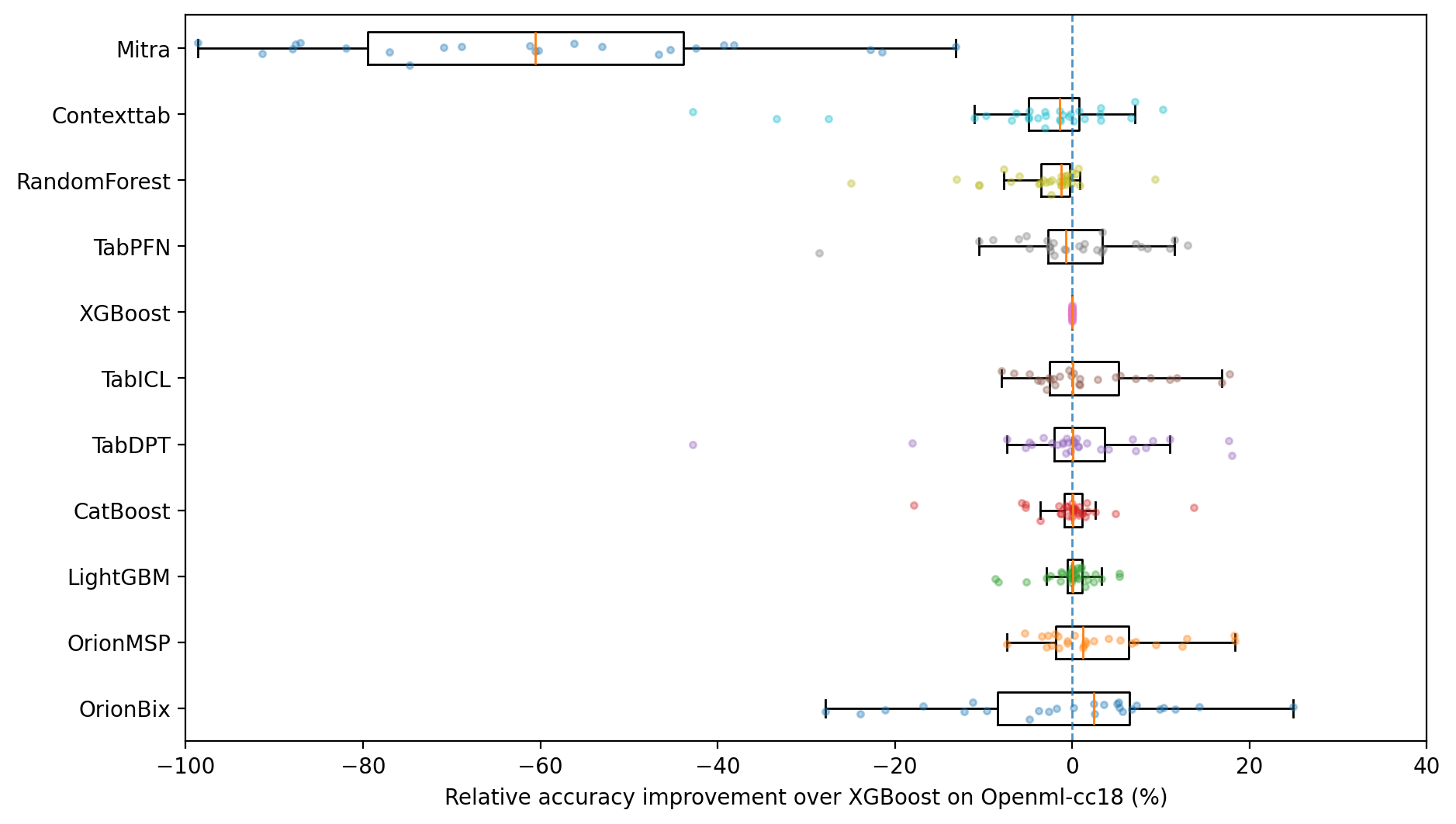}
    \caption{Relative accuracy improvement over XGBoost on OPENML-CC18 Benchmark}
    \label{fig:imp_openml}
  \end{subfigure}
  \caption{Relative accuracy improvement over XGBoost on three benchmarks.}
  \label{fig:imp}
\end{figure}

Figure \ref{fig:imp} presents the relative accuracy improvement (\%) of each method over XGBoost, evaluated per dataset across the three benchmarks: TALENT (Figure \ref{fig:imp_talent}), TabZilla (Figure \ref{fig:imp_tabzilla}), and OpenML-cc18 (Figure \ref{fig:imp_openml}). Each point corresponds to a per-dataset delta, while boxplots summarize the distribution (median, interquartile range, and whiskers). The dashed vertical line denotes parity with XGBoost (0\%).

Across all three benchmarks, Orion-MSP consistently improves upon the strong XGBoost baseline. On TabZilla and OpenML-cc18, Orion-MSP achieves positive median relative accuracy with a compact interquartile range and few negative outliers, indicating both higher accuracy and greater reliability. On TALENT, Orion-MSP reaches parity in median performance but exhibits lower variance than most neural baselines.

Among classical boosted trees, LightGBM, CatBoost, and Random Forest cluster tightly around parity, showing comparable behavior. In contrast, tabular foundation models (TFMs), notably Mitra and ContextTab, exhibit pronounced negative shifts and high variance. TabPFN and TabICL perform competitively and occasionally outperform Orion-MSP on specific datasets, yet their broader variance and heavier left tails reveal less consistent behavior. Overall, Orion-MSP matches or surpasses their central performance and achieves the best mean–variance trade-off, confirming the benefits of our model design for tabular generalization.

To further quantify cross-dataset performance, we computed per-dataset ranks across all 11 methods (lower is better) for each benchmark, averaged them over datasets, and conducted a one-way Friedman test to assess overall differences. When significant, Nemenyi post-hoc tests were applied, and the resulting critical difference (CD) diagrams were plotted at $\alpha = 0.05$; methods connected by the CD bar are not significantly different.

\begin{figure}[htbp]
  \centering
  \begin{subfigure}[b]{0.9\textwidth}
    \centering
    \includegraphics[width=\textwidth]{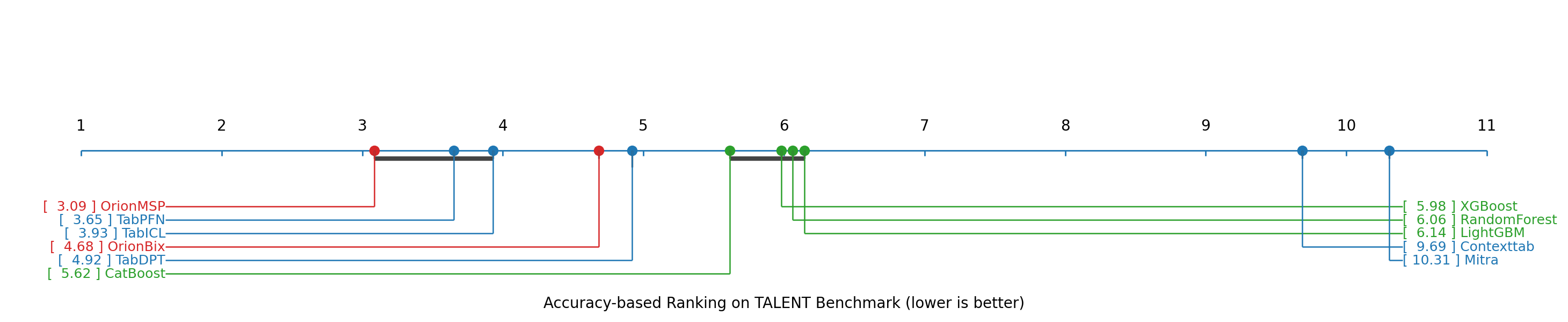}
    \caption{Relative accuracy improvement over XGBoost on TALENT Benchmark}
    \label{fig:rank_talent}
  \end{subfigure}
  \hfill
  \begin{subfigure}[b]{0.9\textwidth}
    \centering
    \includegraphics[width=\textwidth]{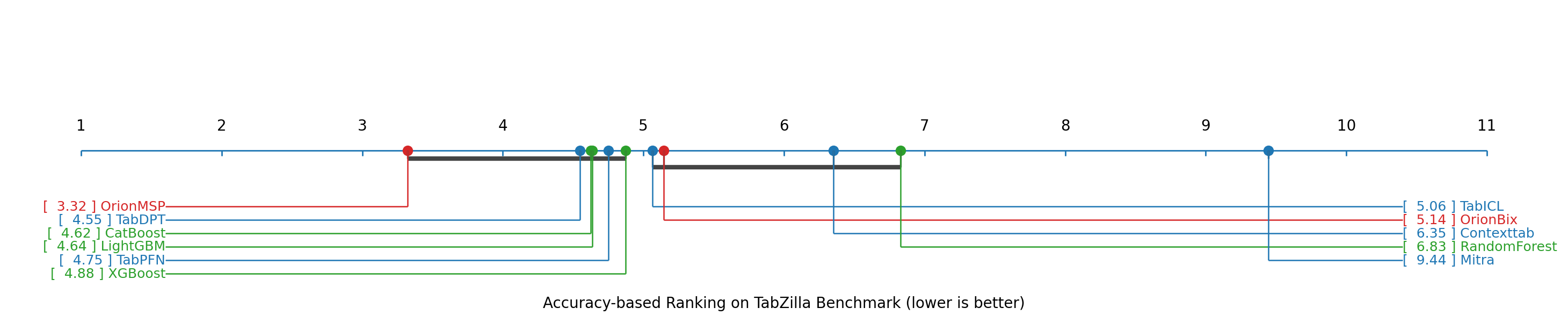}
    \caption{Relative accuracy improvement over XGBoost on TabZilla Benchmark}
    \label{fig:rank_tabzilla}
  \end{subfigure}
   \hfill
  \begin{subfigure}[b]{0.9\textwidth}
    \centering
    \includegraphics[width=\textwidth]{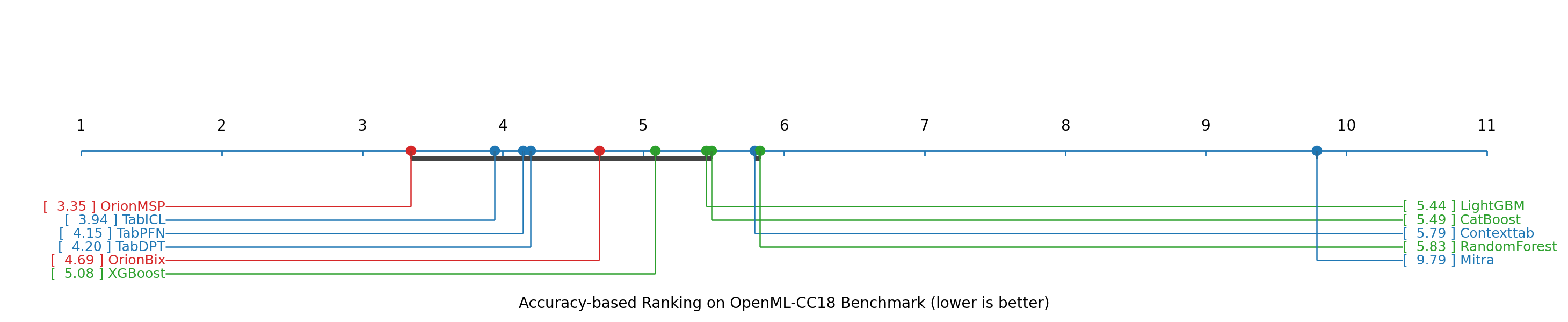}
    \caption{Relative accuracy improvement over XGBoost on OPENML-CC18 Benchmark}
    \label{fig:rank_openml}
  \end{subfigure}
  \caption{}
  \label{fig:cd}
\end{figure}

As shown in Figure \ref{fig:cd}, Orion-MSP attains the best average rank on all three benchmarks—3.09 on TALENT, 3.32 on TabZilla, and 3.35 on OpenML-cc18—appearing as the leftmost method in each CD diagram. TabICL, TabPFN, and TabDPT follow closely, typically lagging by $\approx$ 0.5–1.5 rank points. Tree ensembles (XGBoost, LightGBM, CatBoost, Random Forest) occupy a middle tier with average ranks of 4.6–6.2, showing parity among themselves but a consistent gap to Orion-MSP and the stronger TFMs. Finally, ContextTab and Mitra form the rightmost group (ranks $\approx$ 9–10), confirming the underperformance seen in the improvement plots.

In summary, the CD diagrams corroborate our main finding: Orion-MSP is the top performer across diverse tabular benchmarks, outperforming tree ensembles on average and matching or exceeding pretrained tabular foundation models while maintaining a favorable significance profile.

\section{Datasets}
\label{app:data}

\begin{figure}[htbp]
    \centering
    \includegraphics[width=1\linewidth]{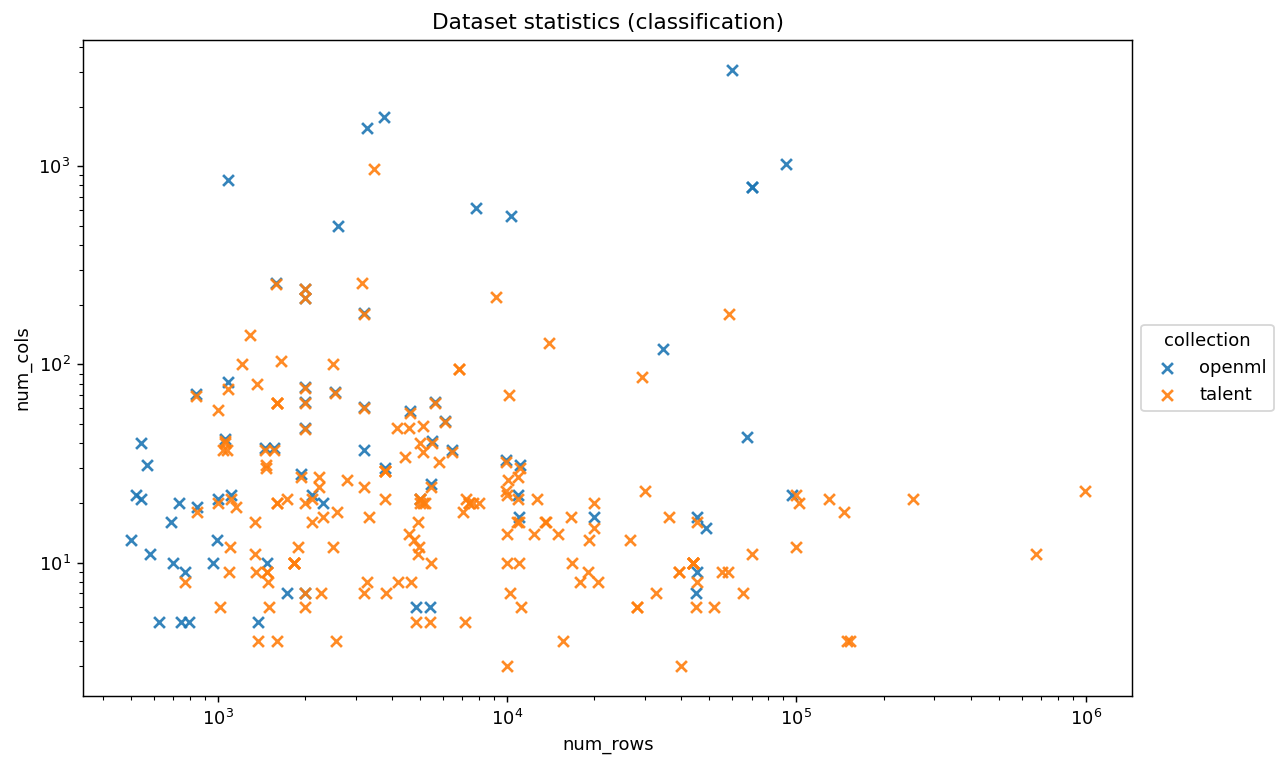}
    \caption{Column and row distribution of the evaluated datasets.}
    \label{fig:benchmarks}
\end{figure}
Full details of all datasets and benchmarks are summarized in below, with their row and column distributions visualized in Figure \ref{fig:benchmarks}.

\subsubsection*{OpenML-CC18 Benchmark Datasets}

Table~\ref{tab:datasets_openml} lists all datasets from the OpenML-CC18 benchmark suite used in our evaluation.

\subsubsection*{TALENT Benchmark Datasets}

Table~\ref{tab:datasets_talent} lists all datasets from the TALENT benchmark suite used in our evaluation.

\subsubsection*{TabZilla Benchmark Datasets}

Table~\ref{tab:datasets_tabzilla} lists all datasets from the TabZilla benchmark suite. TabZilla uses OpenML dataset IDs, and these datasets are specifically selected for evaluating neural network performance on tabular data.

\footnotesize
\begin{longtable}{p{1.2cm}p{4.5cm}p{1.5cm}rrrll}
\caption{OpenML-CC18 benchmark datasets (72 datasets).}
\label{tab:datasets_openml}\\
\toprule
Benchmark & Dataset Name & Domain & Samples & Features & Classes & Task Type & Used In Experimentation \\
\midrule
\endfirsthead

\caption[]{ Details of OpenML-CC18 benchmark datasets.} \\

\toprule
Benchmark & Dataset Name & Domain & Samples & Features & Classes & Task Type & Used In Experimentation \\
\midrule
\endhead

\midrule
\multicolumn{7}{r}{{Continued on next page}} \\
\midrule
\endfoot

\bottomrule
\endlastfoot
   OpenML & OpenML-ID-3 & Other & 3196 & 37 & 2 & binclass & Yes \\
   OpenML & OpenML-ID-6 & Handwriting & 20000 & 17 & 26 & multiclass & No \\
   OpenML & OpenML-ID-11 & Other & 625 & 5 & 3 & multiclass & Yes \\
   OpenML & OpenML-ID-12 & Other & 2000 & 217 & 10 & multiclass & Yes \\
   OpenML & OpenML-ID-14 & Other & 2000 & 77 & 10 & multiclass & Yes\\
   OpenML & OpenML-ID-15 & Healthcare & 699 & 10 & 2 & binclass & Yes \\
   OpenML & OpenML-ID-16 & Other & 2000 & 65 & 10 & multiclass & Yes \\
   OpenML & OpenML-ID-18 & Other & 2000 & 7 & 10 & multiclass & Yes \\
   OpenML & OpenML-ID-22 & Other & 2000 & 48 & 10 & multiclass & Yes\\
   OpenML & OpenML-ID-23 & Healthcare & 1473 & 10 & 3 & multiclass & Yes \\
   OpenML & OpenML-ID-28 & Handwriting & 5620 & 65 & 10 & multiclass & Yes \\
   OpenML & OpenML-ID-29 & Finance & 690 & 16 & 2 & binclass & Yes \\
   OpenML & OpenML-ID-31 & Finance & 1000 & 21 & 2 & binclass & Yes \\
   OpenML & OpenML-ID-32 & Handwriting & 10992 & 17 & 10 & multiclass & Yes \\
   OpenML & OpenML-ID-37 & Healthcare & 768 & 9 & 2 & binclass & Yes \\
   OpenML & OpenML-ID-38 & Healthcare & 3772 & 30 & 2 & binclass & Yes \\
   OpenML & OpenML-ID-44 & Other & 4601 & 58 & 2 & binclass & Yes \\
   OpenML & OpenML-ID-46 & Other & 3190 & 61 & 3 & multiclass & Yes \\
   OpenML & OpenML-ID-50 & Other & 958 & 10 & 2 & binclass & Yes \\
   OpenML & OpenML-ID-54 & Other & 846 & 19 & 4 & multiclass & Yes \\
   OpenML & OpenML-ID-151 & Other & 45312 & 9 & 2 & binclass & Yes \\
   OpenML & OpenML-ID-182 & Other & 6430 & 37 & 6 & multiclass & Yes \\
   OpenML & OpenML-ID-188 & Other & 736 & 20 & 5 & multiclass & Yes \\
   OpenML & OpenML-ID-300 & Other & 7797 & 618 & 26 & multiclass & No \\
   OpenML & OpenML-ID-307 & Other & 990 & 13 & 11 & multiclass & No \\
   OpenML & OpenML-ID-458 & Other & 841 & 71 & 4 & multiclass & Yes \\
   OpenML & OpenML-ID-469 & Healthcare & 797 & 5 & 6 & multiclass & Yes \\
   OpenML & OpenML-ID-554 & Other & 70000 & 785 & 10 & multiclass & Yes \\
   OpenML & OpenML-ID-1049 & Other & 1458 & 38 & 2 & binclass & Yes \\
   OpenML & OpenML-ID-1050 & Other & 1563 & 38 & 2 & binclass & Yes \\
   OpenML & OpenML-ID-1053 & Other & 10885 & 22 & 2 & binclass & Yes \\
   OpenML & OpenML-ID-1063 & Other & 522 & 22 & 2 & binclass & Yes \\
   OpenML & OpenML-ID-1067 & Other & 2109 & 22 & 2 & binclass & Yes \\
   OpenML & OpenML-ID-1068 & Other & 1109 & 22 & 2 & binclass & Yes \\
   OpenML & OpenML-ID-1461 & Finance & 45211 & 17 & 2 & binclass & Yes \\
   OpenML & OpenML-ID-1462 & Finance & 1372 & 5 & 2 & binclass & Yes \\
   OpenML & OpenML-ID-1464 & Healthcare & 748 & 5 & 2 & binclass & Yes \\
   OpenML & OpenML-ID-1468 & Other & 1080 & 857 & 9 & multiclass & Yes \\
   OpenML & OpenML-ID-1475 & Other & 6118 & 52 & 6 & multiclass & Yes \\
   OpenML & OpenML-ID-1478 & Other & 10299 & 562 & 6 & multiclass & Yes \\
   OpenML & OpenML-ID-1480 & Healthcare & 583 & 11 & 2 & binclass & Yes \\
   OpenML & OpenML-ID-1485 & Other & 2600 & 501 & 2 & binclass & Yes \\
   OpenML & OpenML-ID-1486 & Other & 34465 & 119 & 2 & binclass & No \\
   OpenML & OpenML-ID-1487 & Other & 2534 & 73 & 2 & binclass & Yes \\
   OpenML & OpenML-ID-1489 & Other & 5404 & 6 & 2 & binclass & Yes \\
   OpenML & OpenML-ID-1494 & Other & 1055 & 42 & 2 & binclass & Yes \\
   OpenML & OpenML-ID-1497 & Other & 5456 & 25 & 4 & multiclass & Yes \\
   OpenML & OpenML-ID-1501 & Other & 1593 & 257 & 10 & multiclass & Yes \\
   OpenML & OpenML-ID-1510 & Other & 569 & 31 & 2 & binclass & Yes \\
   OpenML & OpenML-ID-1590 & Other & 48842 & 15 & 2 & binclass & Yes \\
   OpenML & OpenML-ID-4134 & Other & 3751 & 1777 & 2 & binclass & No \\
   OpenML & OpenML-ID-4534 & Other & 11055 & 31 & 2 & binclass & Yes \\
   OpenML & OpenML-ID-4538 & Other & 9873 & 33 & 5 & multiclass & Yes \\
   OpenML & OpenML-ID-6332 & Other & 540 & 40 & 2 & binclass & Yes \\
   OpenML & OpenML-ID-23381 & Retail & 500 & 13 & 2 & binclass & Yes\\
   OpenML & OpenML-ID-23517 & Other & 96320 & 22 & 2 & binclass & No \\
   OpenML & OpenML-ID-40499 & Other & 5500 & 41 & 11 & multiclass & No \\
   OpenML & OpenML-ID-40668 & Games & 67557 & 43 & 3 & multiclass & No \\
   OpenML & OpenML-ID-40670 & Other & 3186 & 181 & 3 & multiclass & Yes \\
   OpenML & OpenML-ID-40701 & Other & 5000 & 21 & 2 & binclass & Yes \\
   OpenML & OpenML-ID-40923 & Other & 92000 & 1025 & 46 & multiclass & No \\
   OpenML & OpenML-ID-40927 & Handwriting & 60000 & 3073 & 10 & multiclass & No \\
   OpenML & OpenML-ID-40966 & Other & 1080 & 82 & 8 & multiclass & No \\
   OpenML & OpenML-ID-40975 & Other & 1728 & 7 & 4 & multiclass & Yes \\
   OpenML & OpenML-ID-40978 & Other & 3279 & 1559 & 2 & binclass & Yes \\
   OpenML & OpenML-ID-40979 & Other & 2000 & 241 & 10 & multiclass & Yes \\
   OpenML & OpenML-ID-40982 & Other & 1941 & 28 & 7 & multiclass & Yes \\
   OpenML & OpenML-ID-40983 & Other & 4839 & 6 & 2 & binclass & Yes \\
   OpenML & OpenML-ID-40984 & Other & 2310 & 20 & 7 & multiclass & Yes \\
   OpenML & OpenML-ID-40994 & Other & 540 & 21 & 2 & binclass & Yes \\
   OpenML & OpenML-ID-40996 & Other & 70000 & 785 & 10 & multiclass & No \\
   OpenML & OpenML-ID-41027 & Games & 44819 & 7 & 3 & multiclass & Yes \\

\end{longtable}

\footnotesize
\begin{longtable}{p{1.5cm}p{4.5cm}p{1.5cm}rrrll}
\caption{TALENT benchmark datasets (auto-discovered, multiple domains).}
\label{tab:datasets_talent}\\
\toprule
Benchmark & Dataset Name & Domain & Samples & Features & Classes & Task Type & Used In Experimentation \\
\midrule
\endfirsthead

\caption[]{ Details of TALENT benchmark datasets.} \\

\toprule
Benchmark & Dataset Name & Domain & Samples & Features & Classes & Task Type & Used In Experimentation \\
\midrule
\endhead

\midrule
\multicolumn{7}{r}{{Continued on next page}} \\
\midrule
\endfoot

\bottomrule
\endlastfoot

   TALENT & ASP-POTASSCO-class & Other & 1294 & 141 & 11 & multiclass & Yes \\
   TALENT & Amazon\_employee\_access & Other & 32769 & 7 & 2 & binclass & Yes \\
   TALENT & BLE\_RSSI\_\_Indoor\_localization & Other & 9984 & 3 & 3 & multiclass & Yes \\
   TALENT & BNG(breast-w) & Healthcare & 39366 & 9 & 2 & binclass & Yes \\
   TALENT & BNG(cmc) & Other & 55296 & 9 & 3 & multiclass & Yes \\
   TALENT & BNG(tic-tac-toe) & Other & 39366 & 9 & 2 & binclass & Yes \\
   TALENT & Bank\_Customer\_Churn & Finance & 10000 & 10 & 2 & binclass & Yes \\
   TALENT & Basketball\_c & Retail & 1340 & 11 & 2 & binclass & Yes \\
   TALENT & CDC\_Diabetes\_Health & Healthcare & 253680 & 21 & 2 & binclass & Yes \\
   TALENT & California-Housing-Class & Other & 20640 & 8 & 2 & binclass & No \\
   TALENT & Cardiovascular-Disease & Healthcare & 70000 & 11 & 2 & binclass & Yes \\
   TALENT & Click\_prediction\_small & Other & 39948 & 3 & 2 & binclass & Yes \\
   TALENT & Credit\_c & Finance & 100000 & 22 & 3 & multiclass & Yes \\
   TALENT & Customer\_Personality\_Analysis & Retail & 2240 & 24 & 2 & binclass & Yes \\
   TALENT & DataScience\_Kiva\_Crowdfunding & Other & 671205 & 11 & 4 & multiclass & No \\
   TALENT & Diabetic\_Retinopathy\_Debrecen & Healthcare & 1151 & 19 & 2 & binclass & Yes \\
   TALENT & E-CommereShippingData & Other & 10999 & 10 & 2 & binclass & Yes \\
   TALENT & Employee & Other & 4653 & 8 & 2 & binclass & Yes \\
   TALENT & FICO-HELOC-cleaned & Other & 9871 & 23 & 2 & binclass & Yes \\
   TALENT & FOREX\_audcad-day-High & Finance & 1834 & 10 & 2 & binclass & No \\
   TALENT & FOREX\_audcad-hour-High & Finance & 43825 & 10 & 2 & binclass & No \\
   TALENT & FOREX\_audchf-day-High & Finance & 1833 & 10 & 2 & binclass & No \\
   TALENT & FOREX\_audjpy-day-High & Finance & 1832 & 10 & 2 & binclass & No \\
   TALENT & FOREX\_audjpy-hour-High & Finance & 43825 & 10 & 2 & binclass & No \\
   TALENT & FOREX\_audsgd-hour-High & Finance & 43825 & 10 & 2 & binclass & No \\
   TALENT & FOREX\_audusd-hour-High & Finance & 43825 & 10 & 2 & binclass & No \\
   TALENT & FOREX\_cadjpy-day-High & Finance & 1834 & 10 & 2 & binclass & No\\
   TALENT & FOREX\_cadjpy-hour-High & Finance & 43825 & 10 & 2 & binclass & No\\
   TALENT & Firm-Teacher\_Clave-Direction & Other & 10800 & 16 & 4 & multiclass & Yes \\
   TALENT & Fitness\_Club\_c & Other & 1500 & 6 & 2 & binclass & Yes \\
   TALENT & GAMETES\_Epistasis\_2-Way & Games & 1600 & 20 & 2 & binclass & Yes \\
   TALENT & GAMETES\_Heterogeneity & Games & 1600 & 20 & 2 & binclass & Yes \\
   TALENT & Gender\_Gap\_in\_Spanish & Other & 4746 & 13 & 3 & multiclass & Yes \\
   TALENT & GesturePhaseSegmentation & Other & 9873 & 32 & 5 & multiclass & Yes \\
   TALENT & HR\_Analytics\_Job\_Change & Other & 19158 & 13 & 2 & binclass & Yes \\
   TALENT & IBM\_HR\_Analytics & Other & 1470 & 31 & 2 & binclass & Yes \\
   TALENT & INNHotelsGroup & Other & 36275 & 17 & 2 & binclass & Yes \\
   TALENT & Indian\_pines & Other & 9144 & 220 & 8 & multiclass & Yes \\
   TALENT & JapaneseVowels & Other & 9961 & 14 & 9 & multiclass & Yes \\
   TALENT & KDDCup09\_upselling & Other & 5128 & 49 & 2 & binclass & Yes \\
   TALENT & MIC & Other & 1649 & 104 & 2 & binclass & Yes \\
   TALENT & MagicTelescope & Other & 19020 & 9 & 2 & binclass & Yes\\
   TALENT & Marketing\_Campaign & Finance & 2240 & 27 & 2 & binclass & Yes\\
   TALENT & Mobile\_Price\_Classification & Telcom & 2000 & 20 & 4 & multiclass & Yes\\
   TALENT & National\_Health\_and\_Nutrition & Healthcare & 2278 & 7 & 2 & binclass & Yes\\
   TALENT & PhishingWebsites & Other & 11055 & 30 & 2 & binclass & Yes \\
   TALENT & PieChart3 & Other & 1077 & 37 & 2 & binclass & Yes \\
   TALENT & Pima\_Indians\_Diabetes & Healthcare & 768 & 8 & 2 & binclass & Yes\\
   TALENT & PizzaCutter3 & Other & 1043 & 37 & 2 & binclass & Yes\\
   TALENT & Pumpkin\_Seeds & Other & 2500 & 12 & 2 & binclass & Yes\\
   TALENT & QSAR\_biodegradation & Healthcare & 1054 & 41 & 2 & binclass & Yes\\
   TALENT & Rain\_in\_Australia & Other & 145460 & 18 & 3 & multiclass & No \\
   TALENT & SDSS17 & Other & 100000 & 12 & 3 & multiclass & Yes\\
   TALENT & Satellite & Other & 5100 & 36 & 2 & binclass & Yes\\
   TALENT & Smoking\_and\_Drinking & Other & 991346 & 23 & 2 & binclass & No\\
   TALENT & Telecom\_Churn\_Dataset & Telcom & 3333 & 17 & 2 & binclass & Yes\\
   TALENT & UJI\_Pen\_Characters & Other & 1364 & 80 & 35 & multiclass & Yes\\
   TALENT & Water\_Quality\_and\_Potability & Manufacturing & 3276 & 8 & 2 & binclass & Yes\\
   TALENT & Wilt & Other & 4821 & 5 & 2 & binclass & Yes\\
   TALENT & abalone & Other & 4177 & 8 & 3 & multiclass & Yes \\
   TALENT & accelerometer & Other & 153004 & 4 & 4 & multiclass & No \\
   TALENT & ada & Other & 4147 & 48 & 2 & binclass & Yes\\
   TALENT & ada\_agnostic & Other & 4562 & 48 & 2 & binclass & Yes\\
   TALENT & ada\_prior & Other & 4562 & 14 & 2 & binclass & Yes\\
   TALENT & airlines\_seed\_0\_nrows & Telcom & 2000 & 7 & 2 & binclass & Yes \\
   TALENT & allbp & Other & 3772 & 29 & 3 & multiclass & Yes\\
   TALENT & allrep & Other & 3772 & 29 & 4 & multiclass & Yes\\
   TALENT & analcatdata\_authorship & Other & 841 & 69 & 4 & multiclass & Yes\\
   TALENT & artificial-characters & Other & 10218 & 7 & 10 & multiclass & Yes\\
   TALENT & autoUniv-au4-2500 & Other & 2500 & 100 & 3 & multiclass & Yes\\
   TALENT & autoUniv-au7-1100 & Other & 1100 & 12 & 5 & multiclass & Yes\\
   TALENT & bank & Finance & 45211 & 16 & 2 & binclass & Yes\\
   TALENT & banknote\_authentication & Finance & 1372 & 4 & 2 & binclass & Yes\\
   TALENT & baseball & Other & 1340 & 16 & 3 & multiclass & Yes\\
   TALENT & car-evaluation & Other & 1728 & 21 & 4 & multiclass & Yes\\
   TALENT & churn & Finance & 5000 & 20 & 2 & binclass & Yes \\
   TALENT & cmc & Other & 1473 & 9 & 3 & multiclass & Yes \\
   TALENT & company\_bankruptcy\_prediction & Finance & 6819 & 95 & 2 & binclass & Yes \\
   TALENT & compass & Other & 16644 & 17 & 2 & binclass & Yes \\
   TALENT & contraceptive\_method\_choice & Other & 1473 & 9 & 3 & multiclass & Yes\\
   TALENT & credit & Finance & 16714 & 10 & 2 & binclass & Yes\\
   TALENT & customer\_satisfaction\_in\_airline & Retail & 129880 & 21 & 2 & binclass & No\\
   TALENT & dabetes\_130-us\_hospitals & Healthcare & 101766 & 20 & 2 & binclass & Yes\\
   TALENT & default\_of\_credit\_card\_clients & Finance & 30000 & 23 & 2 & binclass & Yes\\
   TALENT & delta\_ailerons & Other & 7129 & 5 & 2 & binclass & Yes\\
   TALENT & dis & Other & 3772 & 29 & 2 & binclass & Yes\\
   TALENT & dna & Healthcare & 3186 & 180 & 3 & multiclass & Yes\\
   TALENT & drug\_consumption & Healthcare & 1884 & 12 & 7 & multiclass & Yes\\
   TALENT & dry\_bean\_dataset & Other & 13611 & 16 & 7 & multiclass & Yes\\
   TALENT & eeg-eye-state & Other & 14980 & 14 & 2 & binclass & Yes\\
   TALENT & electricity & Manufacturing & 45312 & 8 & 2 & binclass & Yes\\
   TALENT & estimation\_of\_obesity\_levels & Other & 2111 & 16 & 7 & multiclass & Yes\\
   TALENT & eye\_movements & Healthcare & 10936 & 27 & 3 & multiclass & Yes\\
   TALENT & eye\_movements\_bin & Healthcare & 7608 & 20 & 2 & binclass & Yes\\
   TALENT & first-order-theorem-proving & Other & 6118 & 51 & 6 & multiclass & Yes\\
   TALENT & gas-drift & Other & 13910 & 128 & 6 & multiclass & Yes\\
   TALENT & gina\_agnostic & Other & 3468 & 970 & 2 & binclass & Yes\\
   TALENT & golf\_play\_dataset\_extended & Other & 1095 & 9 & 2 & binclass & Yes\\
   TALENT & heloc & Other & 10000 & 22 & 2 & binclass & Yes\\
   TALENT & hill-valley & Other & 1212 & 100 & 2 & binclass & No\\
   TALENT & house\_16H & Other & 13488 & 16 & 2 & binclass & Yes\\
   TALENT & htru & Other & 17898 & 8 & 2 & binclass & Yes\\
   TALENT & ibm-employee-performance & Other & 1470 & 30 & 2 & binclass & Yes\\
   TALENT & in\_vehicle\_coupon\_recos & Retail & 12684 & 21 & 2 & binclass & Yes\\
   TALENT & internet\_firewall & Other & 65532 & 7 & 4 & multiclass & Yes\\
   TALENT & internet\_usage & Other & 10108 & 70 & 46 & multiclass & Yes\\
   TALENT & jm1 & Other & 10885 & 21 & 2 & binclass & No\\
   TALENT & jungle\_chess\_2pcs\_raw & Other & 44819 & 6 & 3 & multiclass & Yes\\
   TALENT & kc1 & Other & 2109 & 21 & 2 & binclass & No\\
   TALENT & kdd\_ipums\_la\_97-small & Other & 5188 & 20 & 2 & binclass & Yes\\
   TALENT & kr-vs-k & Other & 28056 & 6 & 18 & multiclass & Yes \\
   TALENT & kropt & Other & 28056 & 6 & 18 & multiclass & Yes\\
   TALENT & led24 & Other & 3200 & 24 & 10 & multiclass & Yes \\
   TALENT & led7 & Other & 3200 & 7 & 10 & multiclass & Yes \\
   TALENT & letter & Other & 20000 & 15 & 26 & multiclass & Yes \\
   TALENT & madeline & Other & 3140 & 259 & 2 & binclass & Yes\\
   TALENT & mammography & Other & 11183 & 6 & 2 & binclass & Yes\\
   TALENT & maternal\_health\_risk & Healthcare & 1014 & 6 & 3 & multiclass & Yes\\
   TALENT & mfeat-factors & Other & 2000 & 216 & 10 & multiclass & Yes \\
   TALENT & mfeat-fourier & Other & 2000 & 76 & 10 & multiclass & Yes\\
   TALENT & mfeat-karhunen & Other & 2000 & 64 & 10 & multiclass & Yes \\
   TALENT & mfeat-morphological & Other & 2000 & 6 & 10 & multiclass & Yes\\
   TALENT & mfeat-pixel & Other & 2000 & 240 & 10 & multiclass & Yes\\
   TALENT & mfeat-zernike & Other & 2000 & 47 & 10 & multiclass & Yes \\
   TALENT & mice\_protein\_expression & Other & 1080 & 75 & 8 & multiclass & Yes\\
   TALENT & microaggregation2 & Other & 20000 & 20 & 5 & multiclass & Yes\\
   TALENT & mobile\_c36\_oversampling & Telcom & 51760 & 6 & 2 & binclass & Yes\\
   TALENT & mozilla4 & Other & 15545 & 4 & 2 & binclass & Yes\\
   TALENT & naticusdroid+android & Healthcare & 29332 & 86 & 2 & binclass & Yes\\
   TALENT & national-longitudinal-survey-binary & Other & 4908 & 16 & 2 & binclass & Yes\\
   TALENT & okcupid\_stem & Other & 26677 & 13 & 3 & multiclass & Yes\\
   TALENT & one-hundred-plants-margin & Other & 1600 & 64 & 100 & multiclass & Yes\\
   TALENT & one-hundred-plants-shape & Other & 1600 & 64 & 100 & multiclass & Yes\\
   TALENT & one-hundred-plants-texture & Other & 1599 & 64 & 100 & multiclass & Yes\\
   TALENT & online\_shoppers & Retail & 12330 & 14 & 2 & binclass & No\\
   TALENT & optdigits & Other & 5620 & 64 & 10 & multiclass & Yes\\
   TALENT & ozone-level-8hr & Other & 2534 & 72 & 2 & binclass & Yes\\
   TALENT & page-blocks & Other & 5473 & 10 & 5 & multiclass & Yes\\
   TALENT & pc1 & Other & 1109 & 21 & 2 & binclass & No\\
   TALENT & pc3 & Other & 1563 & 37 & 2 & binclass & No\\
   TALENT & pc4 & Other & 1458 & 37 & 2 & binclass & No\\
   TALENT & pendigits & Other & 10992 & 16 & 10 & multiclass & Yes\\
   TALENT & phoneme & Other & 5404 & 5 & 2 & binclass & Yes\\
   TALENT & pol & Other & 10082 & 26 & 2 & binclass & Yes\\
   TALENT & predict\_students\_dropout & Other & 4424 & 34 & 3 & multiclass & Yes \\
   TALENT & qsar & Other & 1055 & 40 & 2 & binclass & Yes \\
   TALENT & rice\_cammeo\_and\_osmancik & Other & 3810 & 7 & 2 & binclass & Yes\\
   TALENT & ringnorm & Other & 7400 & 20 & 2 & binclass & Yes\\
   TALENT & rl & Other & 4970 & 12 & 2 & binclass & No\\
   TALENT & satimage & Other & 6430 & 36 & 6 & multiclass & Yes\\
   TALENT & segment & Other & 2310 & 17 & 7 & multiclass & Yes\\
   TALENT & seismic+bumps & Other & 2584 & 18 & 2 & binclass & Yes\\
   TALENT & semeion & Other & 1593 & 256 & 10 & multiclass & No\\
   TALENT & shuttle & Other & 58000 & 9 & 7 & multiclass & Yes\\
   TALENT & spambase & Other & 4601 & 57 & 2 & binclass & Yes\\
   TALENT & splice & Other & 3190 & 60 & 3 & multiclass & Yes\\
   TALENT & sports\_articles\_for\_objectivity & Other & 1000 & 59 & 2 & binclass & Yes\\
   TALENT & statlog & Other & 1000 & 20 & 2 & binclass& Yes \\
   TALENT & steel\_plates\_faults & Other & 1941 & 27 & 7 & multiclass & Yes\\
   TALENT & sylvine & Other & 5124 & 20 & 2 & binclass & Yes\\
   TALENT & taiwanese\_bankruptcy & Finance & 6819 & 95 & 2 & binclass & Yes\\
   TALENT & telco-customer-churn & Telcom & 7043 & 18 & 2 & binclass & Yes\\
   TALENT & texture & Other & 5500 & 40 & 11 & multiclass & Yes\\
   TALENT & thyroid & Healthcare & 7200 & 21 & 3 & multiclass & Yes\\
   TALENT & thyroid-ann & Healthcare & 3772 & 21 & 3 & multiclass & Yes\\
   TALENT & thyroid-dis & Healthcare & 2800 & 26 & 5 & multiclass & Yes\\
   TALENT & turiye\_student\_evaluation & Other & 5820 & 32 & 5 & multiclass & Yes\\
   TALENT & twonorm & Other & 7400 & 20 & 2 & binclass & Yes\\
   TALENT & vehicle & Other & 846 & 18 & 4 & multiclass & Yes\\
   TALENT & volkert & Other & 58310 & 180 & 10 & multiclass & Yes \\
   TALENT & walking-activity & Other & 149332 & 4 & 22 & multiclass & No \\
   TALENT & wall-robot-navigation & Other & 5456 & 24 & 4 & multiclass & Yes\\
   TALENT & water\_quality & Manufacturing & 7996 & 20 & 2 & binclass & Yes\\
   TALENT & waveform-5000 & Other & 5000 & 40 & 3 & multiclass & Yes\\
   TALENT & waveform\_database\_generator & Other & 4999 & 21 & 3 & multiclass & Yes\\
   TALENT & waveform\_database\_generator-v2 & Other & 5000 & 21 & 3 & multiclass & Yes\\
   TALENT & website\_phishing & Other & 1353 & 9 & 3 & multiclass & Yes\\
   TALENT & wine & Manufacturing & 2554 & 4 & 2 & binclass & No \\
   TALENT & wine-quality-red & Manufacturing & 1599 & 4 & 6 & multiclass & Yes \\
   TALENT & wine-quality-white & Manufacturing & 4898 & 11 & 7 & multiclass & Yes \\
   TALENT & yeast & Other & 1484 & 8 & 10 & multiclass & Yes\\

\end{longtable}

\footnotesize
\begin{longtable}{p{1.5cm}p{3.5cm}p{2.5cm}rrrll}
\caption{TabZilla benchmark datasets (36 datasets via OpenML).}
\label{tab:datasets_tabzilla}\\
\toprule
Benchmark & Dataset Name & Domain & Samples & Features & Classes & Task Type & Used In Experimentation \\
\midrule
\endfirsthead

\caption[]{ Details of TabZilla benchmark datasets.} \\

\toprule
Benchmark & Dataset Name & Domain & Samples & Features & Classes & Task Type & Used In Experimentation\\
\midrule
\endhead

\midrule
\multicolumn{7}{r}{{Continued on next page}} \\
\midrule
\endfoot

\bottomrule
\endlastfoot

   TabZilla & OpenML-ID-999 & Health & 226 & 70 & 2 & binclass & Yes\\
   TabZilla & OpenML-ID-10 & Health & 148 & 19 & 4 & multiclass & Yes\\
   TabZilla & OpenML-ID-11 & Other & 625 & 5 & 3 & multiclass & Yes \\
   TabZilla & OpenML-ID-14 & Other & 2000 & 77 & 10 & multiclass & Yes \\
   TabZilla & OpenML-ID-22 & Other & 2000 & 48 & 10 & multiclass & Yes \\
   TabZilla & OpenML-ID-29 & Finance & 690 & 16 & 2 & binclass  & Yes\\
   TabZilla & OpenML-ID-27 & Health & 368 & 23 & 2 & binclass & Yes\\
   TabZilla & OpenML-ID-31 & Finance & 1000 & 21 & 2 & binclass & Yes \\
   TabZilla & OpenML-ID-46 & Other & 3190 & 61 & 3 & multiclass & Yes \\
   TabZilla & OpenML-ID-54 & Other & 846 & 19 & 4 & multiclass & Yes \\
   TabZilla & OpenML-ID-333 & Other & 556 & 7 & 2 & binclass & Yes \\
   TabZilla & OpenML-ID-1067 & Other & 2109 & 22 & 2 & binclass & Yes \\
   TabZilla & OpenML-ID-1468 & Other & 1080 & 857 & 9 & multiclass & Yes \\
   TabZilla & OpenML-ID-1494 & Other & 1055 & 42 & 2 & binclass & Yes \\
   TabZilla & OpenML-ID-43973 & Other & 3172 & 6 & 2 & binclass & Yes \\
   TabZilla & OpenML-ID-1043 & Other & 4562 & 49 & 2 & binclass & Yes \\
   TabZilla & OpenML-ID-43945 & Other & 38474 & 9 & 2 & binclass & Yes\\
   TabZilla & OpenML-ID-1486 & Other & 34465 & 119 & 2 & binclass & No\\
   TabZilla & OpenML-ID-42825 & Other & 8378 & 123 & -- & -- & No\\
   TabZilla & OpenML-ID-4538 & Other & 9873 & 33 & 5 & multiclass & Yes \\
   TabZilla & OpenML-ID-23512 & Other & 98050 & 29 & 2 & binclass & No\\
   TabZilla & OpenML-ID-4134 & Other & 3751 & 1777 & 2 & binclass & Yes\\
   TabZilla & OpenML-ID-470 & Other & 672 & 10 & 2 & binclass & No\\
   TabZilla & OpenML-ID-1493 & Other & 1599 & 65 & 100 & multiclass & Yes\\
   TabZilla & OpenML-ID-1459 & Other & 10218 & 8 & 10 & multiclass & Yes\\
   TabZilla & OpenML-ID-41027 & Games & 44819 & 7 & 3 & multiclass & Yes\\
   TabZilla & OpenML-ID-40981 & Other & 690 & 15 & 2 & binclass & Yes\\
   TabZilla & OpenML-ID-934 & Other & 1156 & 6 & 2 & binclass & Yes\\
   TabZilla & OpenML-ID-1565 & Health & 294 & 14 & 5 & multiclass & Yes\\
   TabZilla & OpenML-ID-41150 & Other & 130064 & 51 & 2 & binclass & No\\
   TabZilla & OpenML-ID-41159 & Other & 20000 & 4297 & 2 & binclass & No\\
   TabZilla & OpenML-ID-846 & Other & 16599 & 19 & 2 & binclass & Yes \\
   TabZilla & OpenML-ID-1169 & Other & 539383 & 8 & 2 & binclass & No\\
   TabZilla & OpenML-ID-41147 & Other & 425240 & 79 & 2 & binclass & Yes\\
   TabZilla & OpenML-ID-41143 & Other & 2984 & 145 & 2 & binclass & Yes\\
   TabZilla & OpenML-ID-1567 & Other & 1025009 & 11 & 10 & multiclass & No\\

\end{longtable}

\end{document}